\documentclass[sigconf,nonacm]{acmart}
\setcopyright{none}
\usepackage{amsmath}
\usepackage{booktabs,multirow,graphicx,xcolor,colortbl,array}
\usepackage{subcaption}
\usepackage{placeins}
\graphicspath{{figures/}}
\definecolor{catTrad}{HTML}{E7F1FC}
\definecolor{catGen}{HTML}{FFF2DC}
\definecolor{catSID}{HTML}{E9F3E6}
\definecolor{serveRow}{HTML}{F5DCD8}
\definecolor{bestRow}{HTML}{F5DCD8}
\definecolor{secondCell}{HTML}{ECEDEF}
\definecolor{ablHServeTag}{HTML}{DCEAF8}
\definecolor{ablDATag}{HTML}{F9E9C9}
\definecolor{ablRule}{HTML}{DDE2E9}
\newsavebox{\tableShadeMeasure}
\newsavebox{\tableShadeNormal}
\newsavebox{\tableShadeVisible}
\newcommand{\tableShadePaint}[3]{%
  \rlap{\smash{\kern#2\relax\color{#1}%
    \vrule width\dimexpr#3\relax height6pt depth2pt}}}
\providecommand{\second}[1]{%
  \begingroup
  \sbox{\tableShadeMeasure}{\colorbox{secondCell}{#1}}%
  \leavevmode
  \tableShadePaint{secondCell}{0pt}{\wd\tableShadeMeasure}%
  \rlap{\hspace{\fboxsep}#1}%
  \phantom{\usebox{\tableShadeMeasure}}%
  \endgroup}
\newcommand{\redcell}[1]{%
  \begingroup
  \sbox{\tableShadeVisible}{#1}%
  \sbox{\tableShadeNormal}{{\mdseries\renewcommand{\textbf}[1]{##1}#1}}%
  \leavevmode
  \tableShadePaint{serveRow}%
    {\dimexpr(\wd\tableShadeVisible-\wd\tableShadeNormal)/2-\fboxsep\relax}%
    {\dimexpr\wd\tableShadeNormal+2\fboxsep\relax}%
  #1%
  \endgroup}
\newcommand{\categoryspan}[4]{%
  \makebox[78pt][l]{\smash{\raisebox{#3}{%
    \begingroup\setlength{\fboxsep}{0pt}%
    \colorbox{#1}{\parbox[c][#2][c]{78pt}{\hspace*{3pt}%
      \parbox{72pt}{\fontsize{6.7}{7.6}\selectfont\bfseries #4}}}%
    \endgroup}}}}
\providecommand{\catrect}[3]{\colorbox{#1}{\parbox[c]{72pt}{\fontsize{6.7}{7.6}\selectfont\bfseries #3}}}
\providecommand{\ablatag}[2]{\colorbox{#1}{\makebox[88pt][l]{\fontsize{6.8}{7.8}\selectfont #2}}}

\newcommand{\HCDA}{\mathcal L_{\mathrm{HCDA}}}
\newcommand{\RIGD}{\mathcal L_{\mathrm{RIGD}}}
\newcommand{\softplus}{\operatorname{softplus}}
\newcommand{\stopgrad}{\operatorname{stopgrad}}

\begin{document}
\title{MuSID-KD: Multigranularity Knowledge Distillation for Generative Recommendation with Semantic IDs}
\author{Fengshi Zhang}
\authornote{Fengshi Zhang and Xiyan Ding contributed equally to this work.}
\affiliation{%
  \institution{East China Normal University}
  \city{Shanghai}
  \country{China}
}

\author{Xiyan Ding}
\authornotemark[1]
\affiliation{%
  \institution{East China Normal University}
  \city{Shanghai}
  \country{China}
}

\author{Shuxiao Xie}
\affiliation{%
  \institution{Fudan University}
  \city{Shanghai}
  \country{China}
}

\author{Wei Zhang}
\authornote{Corresponding author.}
\affiliation{%
  \institution{East China Normal University}
  \city{Shanghai}
  \country{China}
}

\begin{abstract}
Generative recommendation with semantic IDs (SIDs) represents each item as a sequence of discrete tokens and autoregressively predicts the next item's SID. Despite strong recommendation performance, repeated decoding over multiple SID positions incurs substantial inference cost. Knowledge distillation transfers knowledge from a powerful generative teacher to a lightweight student, yet conventional objectives do not fully account for multi-token SID generation. We identify two limitations: item-level distribution alignment may not adequately preserve the teacher's ranking of candidate groups sharing SID prefixes, while uniform alignment across SID positions may affect final item ranking differently. We propose \textbf{MuSID-KD}, which combines Hierarchical Candidate Distribution Alignment (HCDA) and Ranking Impact Guided Levelwise Distillation (RIGD). HCDA aggregates item probabilities according to shared SID prefixes and aligns candidate distributions at successive levels, from candidate groups to individual items. RIGD evaluates the reduction in pairwise ranking surrogate loss when one student score contribution is replaced with the teacher's contribution, and allocates supervision accordingly. Experiments on real-world recommendation datasets using three lightweight student architectures demonstrate improved recommendation accuracy and better alignment of prefix marginal distributions and pairwise item rankings, without modifying the student's online inference architecture. Code is available at \url{https://anonymous.4open.science/r/m7k2-review-83f9/}.
\end{abstract}
\maketitle
\noindent\textbf{Relevance to the Web.} Personalized Web recommendation requires both high ranking quality and efficient serving across large item catalogs. MuSID-KD addresses this quality and efficiency tradeoff through knowledge transfer during training while preserving lightweight SID generation at inference.

\section{Introduction}
The growing scale of item catalogs and diversity of user interests have made item representation and preference modeling central problems in recommendation. Conventional recommenders often use independent atomic IDs that do not explicitly encode semantic relationships among items. Generative recommendation with semantic IDs (SIDs) instead represents each item as a sequence of discrete tokens and predicts the next item's SID from user interactions~\citep{tiger,handbook}. Shared SID prefixes organize items within a structured token space, and models such as TIGER have demonstrated the effectiveness of this paradigm~\citep{tiger}.

Autoregressive SID generation nevertheless incurs serving costs: constrained beam search repeatedly expands candidate prefixes and evaluates next token distributions. Knowledge distillation (KD) transfers predictions from a strong teacher to a compact student to reduce this burden~\citep{hinton,rankingdistill}. For example, SID-MLP replaces costly Transformer decoding with lightweight, prefix-conditioned prediction heads while preserving dependencies on the user context and previously generated SID tokens~\citep{sidmlp}. Recently, SID-specific distillation methods such as SmartGR have further exploited SID hierarchy and ranking information~\citep{smartgr}. However, existing methods still leave two supervision gaps in multi-token SID generation:

\begin{figure*}[!t]
\centering
\includegraphics[width=.95\textwidth]{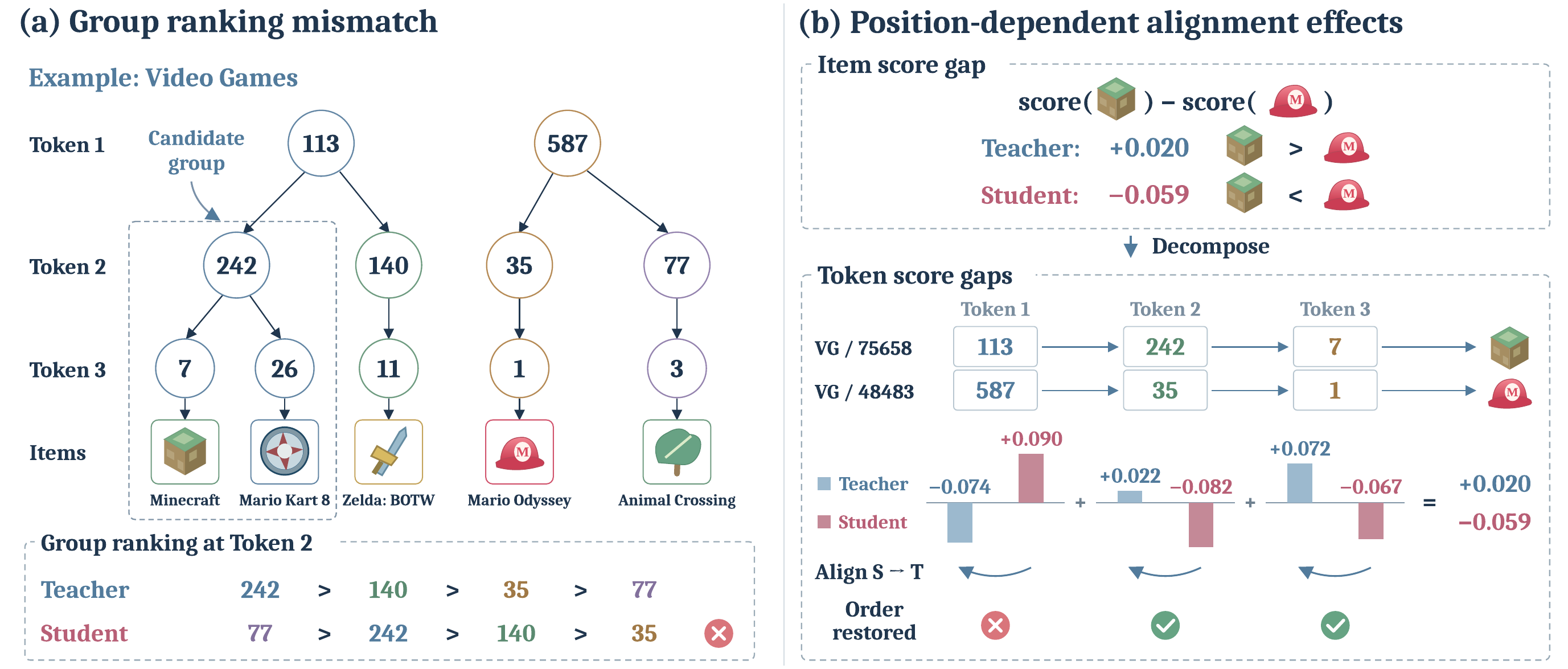}
\Description{Two panel diagram comparing teacher and student candidate group orderings and score contributions at individual SID positions of a pair of video game items.}
\caption{Two supervision problems in distillation with SIDs, illustrated by teacher and student predictions. (a) Teacher and student may rank candidate groups formed by shared SID prefixes differently. (b) For a fixed item pair, differences between teacher and student pairwise score contributions may have opposite signs across SID positions.}
\label{fig:motivation}
\end{figure*}

\begin{itemize}
\setlength{\topsep}{3pt}
\setlength{\partopsep}{0pt}
\setlength{\itemsep}{2pt}
\setlength{\parsep}{0pt}
\item \textbf{Insufficient supervision of candidate distributions across SID granularities.}
Items with a common SID prefix cannot be distinguished until later tokens are generated. At SID position $d$, we call candidate items sharing their first $d$ SID tokens a \emph{candidate group}. Shorter prefixes form coarser groups, while a full SID identifies an individual item. In Figure~\ref{fig:motivation}(a), Minecraft and Mario Kart 8 belong to the group sharing the prefix $(113,242)$. Thus, the teacher's distribution over candidate items also induces coarser prefix marginal distributions over candidate groups.

Item distribution alignment matches the teacher's distribution over candidate items, so candidate group rankings receive only indirect supervision. When student capacity is limited, approximate alignment of item probabilities does not ensure that rankings of the corresponding groups are preserved. In our diagnosis, a student using only candidate item distribution alignment for distillation disagrees with the teacher's group ordering in \textbf{53.8\%} of evaluated comparisons on average across datasets and intermediate SID positions (Figure~\ref{fig:misranking}). Figure~\ref{fig:motivation}(a) illustrates a case where the student ranks group 77 above the teacher's preferred group 242. This motivates explicit alignment of candidate group distributions as well as individual item distributions.

\item \textbf{Uniform supervision across SID positions.}
An item score sums conditional token log probabilities across SID positions. Thus, the gap between two item scores sums the contributions from individual positions. Conventional token distribution distillation supervises predictions at each SID position but does not consider whether matching the teacher there improves the final pairwise ranking. We find that \textbf{69.6\%--71.7\%} of evaluated item pairs exhibit score residuals with opposite signs across SID positions (Figure~\ref{fig:residuals}). In Figure~\ref{fig:motivation}(b), the student's contribution at the first position favors the teacher-preferred item ($+0.090$), whereas the teacher's contribution at that position favors the other item ($-0.074$). At the second and third positions, the teacher favors the preferred item, while the student favors the other item. Uniformly pulling all levels toward the teacher can reduce an already favorable contribution while improving the others.

\end{itemize}

Motivated by these findings, we propose \textbf{MuSID-KD}, a distillation framework tailored to multi-token SID generation. Its \textbf{Hierarchical Candidate Distribution Alignment (HCDA)} objective aggregates candidate item probabilities under each shared SID prefix and aligns the resulting prefix marginal distributions at all SID positions. Its \textbf{Ranking Impact Guided Levelwise Distillation (RIGD)} objective measures how replacement at one SID position changes the pairwise ranking surrogate loss; only positions with positive replacement gain receive nonzero weight. Both objectives operate during offline training and leave the student model and its online decoding unchanged.

\noindent The main contributions of this work are summarized as follows:
\begin{itemize}
\setlength{\topsep}{3pt}
\setlength{\partopsep}{0pt}
\setlength{\itemsep}{2pt}
\setlength{\parsep}{0pt}
    \item We identify two key limitations of existing SID distillation: insufficient supervision of candidate group rankings at intermediate SID positions and uniform teacher matching despite its different effects on final item rankings.
    \item We propose MuSID-KD, combining HCDA to align candidate distributions at multiple SID granularities and RIGD to allocate supervision according to the ranking benefit of matching the teacher at each SID position.
    \item Extensive experiments on four real-world datasets with three lightweight student architectures demonstrate improved recommendation accuracy and knowledge transfer, supported by ablation studies, ranking diagnostics, and efficiency analysis.
\end{itemize}

\section{Related Work}
\subsection{Generative Recommendation with Semantic IDs}
Generative recommendation casts next item prediction as sequence generation. TIGER uses residual quantization to assign semantic token sequences to items and generates SIDs autoregressively~\citep{tiger,rqvae}. Other work studies learned tokenization, SID representation, and efficient generation~\citep{itemtoken,parallelsid}. Unlike recommendation based on atomic IDs, SID models expose multiple token positions and shared prefixes, offering a natural basis for candidate group analysis. Repeated decoding and beam search nevertheless impose online inference costs~\citep{sidmlp}.

\vspace{-2pt}
\subsection{Knowledge Distillation for Recommendation}
Knowledge distillation compresses a teacher's predictive behavior into a smaller model~\citep{hinton}. Recommendation specific approaches transfer rankings, preference distributions, or relationships between items. Ranking Distillation (RD) uses ranked teacher outputs~\citep{rankingdistill}; Relaxed Ranking Distillation (RRD) and related methods refine sampled supervision~\citep{rrd}; RCE-KD studies ranking oriented cross entropy objectives~\citep{rcekd}. In generative modeling, sequence distillation and on policy distribution matching transfer autoregressive behavior~\citep{seqkd,gkd,distillm2}. These objectives are informative baselines but do not directly supervise candidate groups at all SID levels or model each level's influence on item comparisons.

\subsection{Distillation for SID Generation}
Recently, knowledge distillation has begun to be applied to SID-based generative recommendation to reduce its online inference cost. SID-MLP~\citep{sidmlp} retains the teacher model's history encoding capability and replaces the complex Transformer decoder with a lightweight prefix-conditioned MLP prediction network, reducing repeated computation during generation and thus lowering inference overhead. SmartGR~\citep{smartgr} further exploits overlapping SID prefixes for hierarchy-aware token distribution distillation and incorporates cumulative prefix-score relationships among beam-search candidates into student training.

Overall, existing methods leverage SID hierarchy and ranking information but still focus on conditional predictions and prefix rankings. They do not directly align candidate distributions across SID granularities or distinguish how individual SID positions affect final item rankings. MuSID-KD addresses both limitations with structured distillation objectives for multi-token SIDs.

\section{Preliminaries}
\subsection{Generative Recommendation with SIDs}
Given the interaction history $x_u$ of user $u$, each catalog item $i$ is mapped to a unique SID of length $L$,
\[\mathbf z_i=(z_{i,1},z_{i,2},\ldots,z_{i,L}).\]
where $z_{i,d}$ is the token at level $d$ and $\mathbf z_{i,<d}$ denotes the prefix containing the first $d-1$ tokens. Model $m\in\{T,S\}$, with $T$ and $S$ denoting the teacher and student, predicts the conditional token probability $p_m(z_{i,d}\mid x_u,\mathbf z_{i,<d})$. The item score is the sum of conditional token log probabilities along its SID sequence:
\begin{equation}\label{eq:score}
s_m(i\mid u)=\sum_{d=1}^{L}\log p_m(z_{i,d}\mid x_u,\mathbf z_{i,<d}).
\end{equation}
Items are ranked by this score and the top $K$ items are returned. We omit the explicit conditioning on $u$ when unambiguous.

\subsection{Distillation Setting}
We distill a pretrained SID teacher into a lightweight student using the same fixed item to SID mapping and valid SID constraints. The teacher remains fixed; only the student is updated. For each training pair $(x_u,i_u^+)$ with observed next item $i_u^+$, the student receives ground truth next item supervision:
\begin{equation}\label{eq:ce}
\mathcal L_{\mathrm{CE}}=-\mathbb E_{(x_u,i_u^+)}\sum_{d=1}^L\log p_S(z_{i_u^+,d}\mid x_u,\mathbf z_{i_u^+,<d}).
\end{equation}
The teacher generates at most $M$ distinct valid candidate items,
\begin{equation}\label{eq:candidates}
\mathcal C_M(u)=\{i_1,\ldots,i_{|\mathcal C_M(u)|}\},\quad |\mathcal C_M(u)|\le M.
\end{equation}
We cache the teacher's candidate item scores and conditional token log probabilities. The student recomputes SID sequence scores on the same candidate set, allowing the two models to induce different scores, rankings, and candidate distributions. Here $M$ is the maximum number of teacher candidate items; $K$ is the recommendation cutoff with $K\le M$. Distillation is applied only offline.

\section{Analyzing Supervision in SID Distillation}
We first assess whether aligning the candidate item distribution preserves rankings among groups sharing SID prefixes. We then examine how matching the teacher at individual SID positions affects item ranking.

\subsection{Limits of Item Distribution Alignment}
For model $m$, normalizing item scores over the common teacher candidate set defines a temperature scaled candidate distribution
\begin{equation}\label{eq:itemdist}
\pi_m(i\mid u)=\frac{\exp(s_m(i\mid u)/\tau)}{\sum_{j\in\mathcal C_M(u)}\exp(s_m(j\mid u)/\tau)},\quad i\in\mathcal C_M(u),
\end{equation}
where $\tau>0$. At SID position $d$, items belong to the same candidate group when their first $d$ SID tokens match. For a length $d$ prefix $p$, its probability mass is the sum of all candidate item probabilities within that group:
\begin{equation}\label{eq:groupdist}
\pi_m^{(d)}(p\mid u)=\sum_{\substack{i\in\mathcal C_M(u)\\\mathbf z_{i,1:d}=p}}\pi_m(i\mid u).
\end{equation}
The candidate group in Figure~\ref{fig:motivation}(a) sharing prefix $(113,242)$ aggregates the probabilities assigned to Minecraft and Mario Kart 8. Short prefixes yield coarse groups that split as $d$ increases. When $d=L$, a unique complete SID identifies an item, so $\pi_m^{(L)}$ coincides with the item distribution.

Exact matching of the candidate item distribution also matches every prefix marginal distribution. In practice, limited student capacity yields only approximate alignment, so group orderings can still differ. To quantify this issue, we train a student with ground truth supervision and a distillation loss that aligns only the candidate item distribution. At each intermediate SID position, we form four valid competing groups sharing the preceding prefix and assess their ranking against the teacher.
\begin{figure}[!t]
\centering\includegraphics[width=.92\columnwidth]{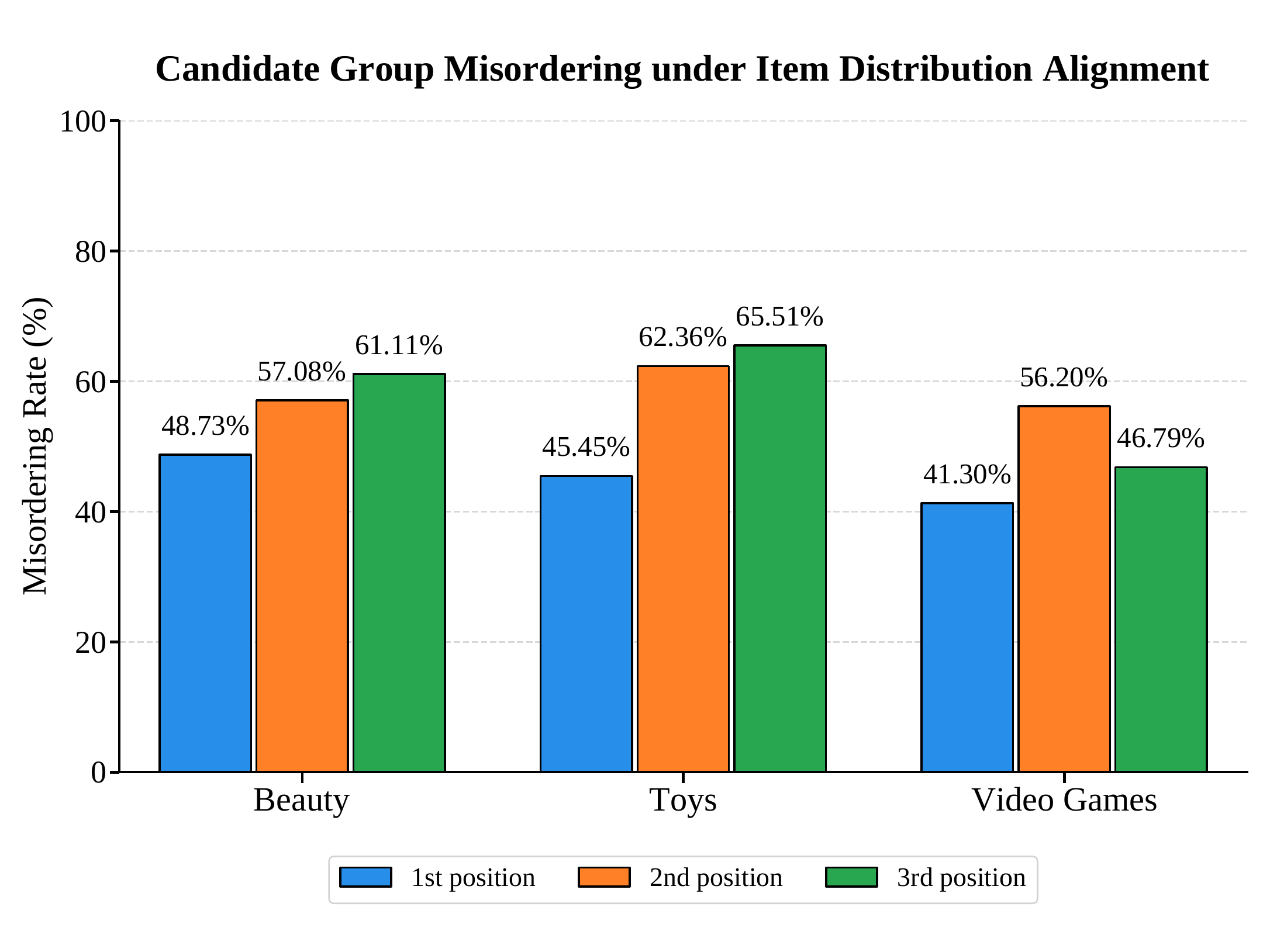}
\Description{Grouped bars reporting candidate group misordering at the first three intermediate SID levels on three recommendation datasets.}
\caption{Candidate group misordering rates at intermediate SID positions when distillation aligns only candidate item probabilities. For each comparison, four groups share their preceding prefix and differ at the current token. A comparison is incorrect when the student reverses any strictly teacher ordered group pair or predicts a tie.}
\label{fig:misranking}
\end{figure}
Figure~\ref{fig:misranking} shows an average misordering rate of \textbf{53.8\%} across the evaluated datasets and intermediate levels.
\par\smallskip\noindent\textbf{Observation 1.}~\emph{Optimizing only the final candidate item distribution does not ensure that a compact student preserves the teacher's candidate group rankings at intermediate SID positions.}
\par\smallskip\noindent Explicit supervision of these distributions is therefore warranted.

\subsection{Different Effects of Teacher Supervision Across SID Positions}
For an item pair $i\succ j$ ordered by the teacher, define the pairwise score contribution at SID position $d$ as the difference between the items' conditional token log probabilities:
\begin{equation}\label{eq:localgap}
\begin{aligned}
g_{m,d}(i,j\mid u)={}&\log p_m(z_{i,d}\mid x_u,\mathbf z_{i,<d})\\
&-\log p_m(z_{j,d}\mid x_u,\mathbf z_{j,<d}).
\end{aligned}
\end{equation}
The pairwise score gap is
\begin{equation}\label{eq:globalgap}
G_m(i,j\mid u)=s_m(i\mid u)-s_m(j\mid u)=\!\sum_{d=1}^L\!g_{m,d}(i,j\mid u).
\end{equation}
If $G_m>0$, model $m$ ranks item $i$ above item $j$. To characterize the direction of matching the teacher at position $d$, define the score residual
\begin{equation}\label{eq:residual}
r_d(i,j\mid u)=g_{T,d}(i,j\mid u)-g_{S,d}(i,j\mid u).
\end{equation}
A positive $r_d$ means that matching the teacher at position $d$ increases the pairwise score gap favoring $i$. A negative $r_d$ means that the same operation decreases the gap. Thus, opposite signs across levels signal different effects on the teacher preferred pair order.

We evaluate pairs spanning the teacher's top $K$ cutoff and count those with both positive and negative score residuals across SID positions.
\begin{figure}[!t]
\centering\includegraphics[width=.97\columnwidth]{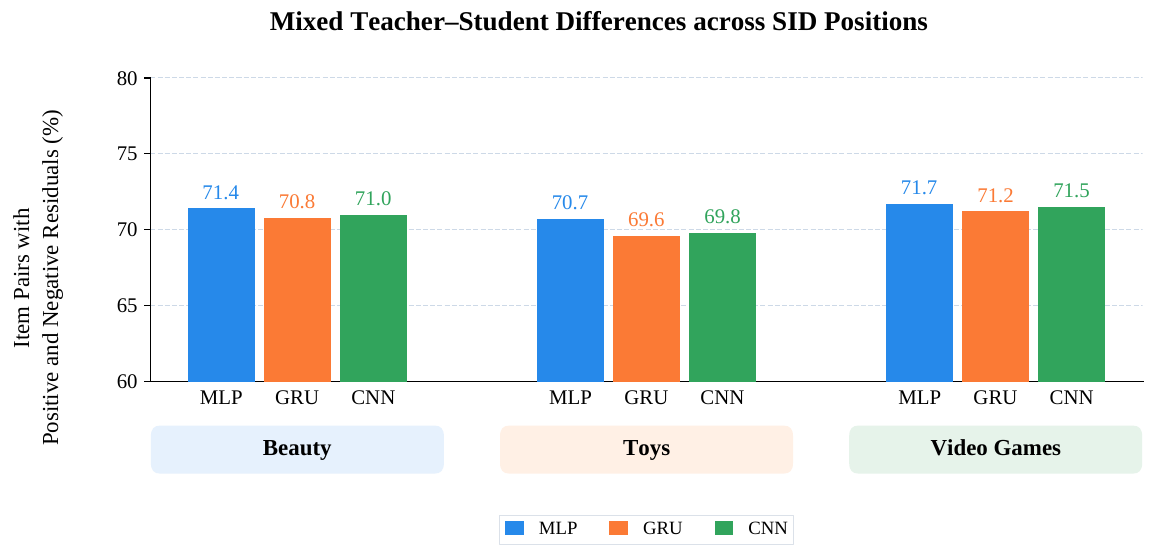}
\Description{Three groups of bars showing the fraction of item pairs with residuals of opposite signs across SID levels.}
\caption{Proportion of item pairs ordered by the teacher whose residual $r_d$ is positive at some SID positions and negative at others, measured for pairs crossing the top $K$ recommendation boundary.}
\label{fig:residuals}
\end{figure}
As Figure~\ref{fig:residuals} shows, this applies to \textbf{69.6\%--71.7\%} of the evaluated pairs.
\par\smallskip\noindent\textbf{Observation 2.}~\emph{Matching the teacher at different SID positions can affect the same final pairwise comparison in opposite directions.}
\par\smallskip\noindent Distillation should therefore assess the effect of each position's score correction rather than supervising all positions uniformly.

\section{MuSID-KD}
\subsection{Framework Overview}
\begin{figure*}[!t]
\centering\includegraphics[width=.98\textwidth]{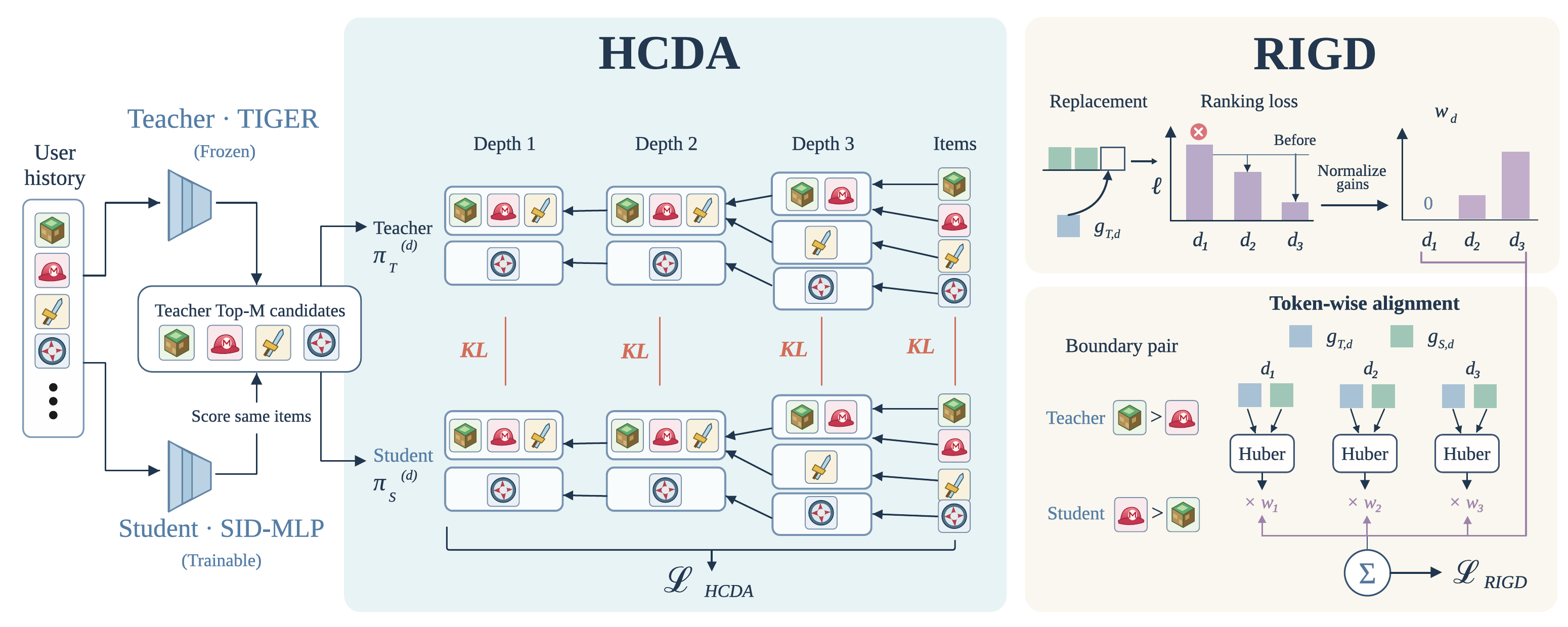}
\Description{Framework with teacher and student candidate generation on the left, hierarchical candidate distribution alignment in the center, and ranking impact guided levelwise distillation on the right.}
\caption{Overview of MuSID-KD. HCDA aligns distributions over candidate groups induced by shared SID prefixes, culminating in the candidate item distribution. RIGD estimates the reduction in pairwise ranking loss when replacing each student score contribution with the teacher's value and uses positive reductions to weight Huber alignment. Both components are absent during online inference.}
\label{fig:framework}
\end{figure*}

Figure~\ref{fig:framework} presents MuSID-KD. The teacher and student score the same cached candidate set. HCDA aligns the candidate item distribution and prefix marginal distributions; RIGD weights score alignment at each SID position according to the reduction in pairwise ranking loss. Both operate only during training.

\subsection{Hierarchical Candidate Distribution Alignment}
Observation 1 motivates supervision over the complete hierarchy of candidate group distributions. We first compute the teacher and student item probabilities in Eq.~\eqref{eq:itemdist}. Next, at each SID position $d$, we aggregate probabilities of candidate items sharing the same length $d$ prefix according to Eq.~\eqref{eq:groupdist}. This gives coherent distributions from coarse candidate groups to individual items. We align them with
\begin{equation}\label{eq:hcda}
\HCDA=\mathbb E_u\left[\sum_{d=1}^L D_{\mathrm{KL}}\!\left(\pi_T^{(d)}(\cdot\mid u)\,\|\,\pi_S^{(d)}(\cdot\mid u)\right)\right].
\end{equation}
Here $D_{\mathrm{KL}}$ is Kullback Leibler divergence. We weight the alignment loss at each SID position equally. Earlier terms align coarse candidate group probability masses; the final term aligns the candidate item distribution. HCDA thus makes the candidate group information that is implicit in item probabilities an explicit training target.

\subsection{Ranking Impact Guided Levelwise Distillation}
Observation 2 motivates deciding \emph{whether and how strongly} to match the teacher's score contribution at each SID position. RIGD considers pairs crossing the teacher's recommendation cutoff. Let $\operatorname{rank}_T(i\mid u)$ be the teacher's rank. Define
\begin{equation}\label{eq:pairs}
\begin{aligned}
\mathcal Q(u)=\{(i,j):&\ \operatorname{rank}_T(i\mid u)\le K,\\
& K<\operatorname{rank}_T(j\mid u)\le B\},\quad K<B\le M.
\end{aligned}
\end{equation}
For each pair, we begin at the first SID position where their tokens differ. Their identical preceding tokens and prefixes have identical conditional probabilities and cannot contribute to the score gap. Let
\begin{equation}\label{eq:eligible}
\begin{aligned}
d_{ij}^{\star}&=\min\{d:z_{i,d}\ne z_{j,d}\},\\
\mathcal D_{ij}&=\{d_{ij}^{\star},d_{ij}^{\star}+1,\ldots,L-1\}.
\end{aligned}
\end{equation}
Only these positions are eligible for RIGD supervision. The final position disambiguates item identity and is excluded from the RIGD loss, but its contribution remains in the pairwise score gap $G_m$. Pairs that first diverge at the final level have no eligible position.

To measure the effect of matching the teacher at position $d$, we keep the other student score contributions fixed and replace only the contribution at $d$ with the teacher's value:
\begin{equation}\label{eq:replace}
\widetilde G_{S,d}(i,j\mid u)=G_S(i,j\mid u)-g_{S,d}(i,j\mid u)+g_{T,d}(i,j\mid u).
\end{equation}
We call this \emph{replacement at a single SID position}. For teacher order $i\succ j$, the margin based Softplus surrogate is
\begin{equation}\label{eq:softplus}
\ell(G)=\softplus(\mu-G),
\end{equation}
where $\mu$ is the ranking margin. A larger gap in the teacher preferred direction yields a lower surrogate loss. We define the nonnegative replacement gain as
\begin{equation}\label{eq:gain}
A_d(i,j\mid u)=\left[\ell(G_S(i,j\mid u))-\ell(\widetilde G_{S,d}(i,j\mid u))\right]_+,
\end{equation}
where $[a]_+=\max(a,0)$. This replacement gain is positive only when matching the teacher at position $d$ reduces the pairwise ranking surrogate loss. We normalize the positive replacement gains within each pair:
\begin{equation}\label{eq:weight}
w_d(i,j\mid u)=\stopgrad\!\left(\frac{A_d(i,j\mid u)}{\max\{\sum_{t\in\mathcal D_{ij}}A_t(i,j\mid u),\epsilon\}}\right).
\end{equation}
The floor $\epsilon>0$ prevents division by zero; $\stopgrad$ blocks gradients through the weights. Positions with zero replacement gain receive zero weight; among the remaining positions, larger gains yield larger weights.

Let $\mathcal Q_{\mathrm{valid}}$ contain the training pairs with nonempty $\mathcal D_{ij}$. RIGD minimizes weighted differences between teacher and student pairwise score contributions:
\begin{equation}\label{eq:rigd}
\begin{aligned}
\RIGD={}&\frac{1}{|\mathcal Q_{\mathrm{valid}}|}\sum_{(u,i,j)\in\mathcal Q_{\mathrm{valid}}}\ \,
\sum_{d\in\mathcal D_{ij}} w_d(i,j\mid u)\\
&\quad\cdot\rho_H\!\left(\frac{g_{S,d}(i,j\mid u)-g_{T,d}(i,j\mid u)}{\sigma_d}\right),
\end{aligned}
\end{equation}
where $\rho_H$ is the Huber loss and $\sigma_d$ is the standard deviation of teacher pairwise score contributions at position $d$, computed from training item pairs. This normalizes score scales across positions. An empty valid set gives zero loss. Replacement at one SID position and the Softplus surrogate are used only to compute the weights; the actual optimization aligns the original contributions with the Huber loss in Eq.~\eqref{eq:rigd}.

\subsection{Training and Inference}
The complete student training objective combines observed next item prediction with the two structured distillation losses:
\begin{equation}\label{eq:objective}
\mathcal L=\mathcal L_{\mathrm{CE}}+\lambda_H\HCDA+\lambda_R\RIGD.
\end{equation}
Both $\lambda_H$ and $\lambda_R$ control the strengths of their corresponding losses. Candidate group aggregation, replacement, and weight computation occur only during offline distillation. Online serving uses the student history encoder, prefix-conditioned token heads, and constrained SID beam search, without additional inference modules.

\section{Experiments}\label{sec:experiments}
We address four questions: \textbf{RQ1}: Does MuSID-KD improve recommendation quality over existing KD methods? \textbf{RQ2}: How do HCDA, RIGD, and their design choices affect performance? \textbf{RQ3}: Do the objectives improve candidate group alignment and item ranking knowledge transfer? \textbf{RQ4}: What training overhead and inference costs result from MuSID-KD?

\begin{table*}[!t]
\centering
\caption{Amazon results (0--1 scale; mean over seeds 42--44). R@K/N@K denote Recall@K/NDCG@K. Gain vs. Base and Gain vs. best are relative percentage improvements over SID-MLP (Base KD) and the strongest other student, respectively, computed from the displayed means; TIGER is a teacher reference and excluded from student comparisons. SD is the standard deviation of MuSID-KD over three seeds. Bold/light-red denote the best student; gray denotes the strongest other student (including ties). VG is Video Games.}
\label{table:main}
\begingroup
\fontsize{6.5}{7.3}\selectfont
\let\tableOneSecond\second
\renewcommand{\second}[1]{\smash{\tableOneSecond{#1}}}
\setlength{\tabcolsep}{1.1pt}
\renewcommand{\arraystretch}{0.90}
\begin{tabular*}{\textwidth}{@{\extracolsep{\fill}}ll*{12}{c}@{}}
\toprule
\textbf{Category} & \textbf{Method} & \multicolumn{4}{c}{\textbf{MLP}} & \multicolumn{4}{c}{\textbf{GRU}} & \multicolumn{4}{c}{\textbf{CNN}} \\
\cmidrule(lr){3-6}\cmidrule(lr){7-10}\cmidrule(l){11-14}
 & & R@5 & N@5 & R@10 & N@10 & R@5 & N@5 & R@10 & N@10 & R@5 & N@5 & R@10 & N@10 \\
\midrule
\multicolumn{14}{@{}l}{\textbf{Beauty}} \\
\cellcolor{white}\catrect{secondCell}{12pt}{Teacher reference} & \rule[-2.5pt]{0pt}{9pt}TIGER (Teacher) & 0.0428 & 0.0283 & 0.0677 & 0.0363 & 0.0428 & 0.0283 & 0.0677 & 0.0363 & 0.0428 & 0.0283 & 0.0677 & 0.0363 \\
\cellcolor{white} & \rule[-2.5pt]{0pt}{9pt}RD & 0.0423 & 0.0274 & 0.0656 & 0.0350 & \second{0.0433} & 0.0284 & \second{0.0671} & \second{0.0362} & 0.0418 & 0.0274 & 0.0647 & 0.0348 \\
\cellcolor{white} & \rule[-2.5pt]{0pt}{9pt}RRD & 0.0431 & 0.0285 & 0.0658 & 0.0354 & 0.0422 & 0.0275 & 0.0663 & 0.0354 & \second{0.0420} & \second{0.0281} & 0.0649 & 0.0352 \\
\cellcolor{white}\categoryspan{catTrad}{26pt}{7.4pt}{Traditional\\Recommendation KD} & \rule[-2.5pt]{0pt}{9pt}RCE-KD & 0.0433 & 0.0280 & 0.0664 & 0.0361 & 0.0431 & 0.0279 & 0.0669 & 0.0356 & 0.0408 & 0.0276 & \second{0.0651} & \second{0.0354} \\
\cellcolor{white} & \rule[-2.5pt]{0pt}{9pt}GKD & 0.0427 & 0.0276 & 0.0666 & 0.0359 & 0.0406 & 0.0273 & 0.0644 & 0.0350 & 0.0388 & 0.0259 & 0.0590 & 0.0323 \\
\cellcolor{white}\categoryspan{catGen}{17pt}{2.9pt}{General AR /\\LLM KD} & \rule[-2.5pt]{0pt}{9pt}DistiLLM-2 & 0.0425 & \second{0.0287} & 0.0662 & \second{0.0363} & 0.0419 & 0.0277 & 0.0659 & 0.0352 & 0.0370 & 0.0251 & 0.0574 & 0.0314 \\
\cellcolor{white} & \rule[-2.5pt]{0pt}{9pt}SID-MLP (Base KD) & 0.0429 & 0.0282 & 0.0660 & 0.0356 & 0.0416 & 0.0281 & 0.0656 & 0.0358 & 0.0380 & 0.0254 & 0.0580 & 0.0319 \\
\cellcolor{white} & \rule[-2.5pt]{0pt}{9pt}SmartGR & \second{0.0435} & 0.0278 & \second{0.0668} & 0.0356 & 0.0426 & \second{0.0286} & 0.0667 & 0.0360 & 0.0373 & 0.0249 & 0.0594 & 0.0321 \\
\cellcolor{white}\categoryspan{catSID}{26pt}{7.4pt}{SID-based generative\\recommendation KD} & \rule[-2.5pt]{0pt}{9pt}\redcell{\textbf{MuSID-KD}} & \redcell{\textbf{0.0442}} & \redcell{\textbf{0.0292}} & \redcell{\textbf{0.0679}} & \redcell{\textbf{0.0369}} & \redcell{\textbf{0.0441}} & \redcell{\textbf{0.0291}} & \redcell{\textbf{0.0682}} & \redcell{\textbf{0.0368}} & \redcell{\textbf{0.0437}} & \redcell{\textbf{0.0288}} & \redcell{\textbf{0.0686}} & \redcell{\textbf{0.0368}} \\
\cmidrule(l){2-14}
\cellcolor{white} & \textit{Gain vs. best (\%)} & +1.61\% & +1.74\% & +1.65\% & +1.65\% & +1.85\% & +1.75\% & +1.64\% & +1.66\% & +4.05\% & +2.49\% & +5.38\% & +3.95\% \\
\cellcolor{white} & \textit{Gain vs. Base (\%)} & +3.03\% & +3.55\% & +2.88\% & +3.65\% & +6.01\% & +3.56\% & +3.96\% & +2.79\% & +15.00\% & +13.39\% & +18.28\% & +15.36\% \\
\cellcolor{white} & SD & 0.00011 & 0.00011 & 0.00025 & 0.00022 & 0.00017 & 0.00011 & 0.00045 & 0.00017 & 0.00020 & 0.00011 & 0.00034 & 0.00011 \\
\midrule
\multicolumn{14}{@{}l}{\textbf{Toys}} \\
\cellcolor{white}\catrect{secondCell}{12pt}{Teacher reference} & \rule[-2.5pt]{0pt}{9pt}TIGER (Teacher) & 0.0405 & 0.0254 & 0.0648 & 0.0332 & 0.0405 & 0.0254 & 0.0648 & 0.0332 & 0.0405 & 0.0254 & 0.0648 & 0.0332 \\
\cellcolor{white} & \rule[-2.5pt]{0pt}{9pt}RD & 0.0379 & 0.0233 & 0.0615 & 0.0309 & 0.0376 & 0.0242 & 0.0615 & 0.0319 & \second{0.0380} & \second{0.0241} & \second{0.0603} & \second{0.0312} \\
\cellcolor{white} & \rule[-2.5pt]{0pt}{9pt}RRD & \second{0.0397} & \second{0.0246} & 0.0624 & 0.0319 & 0.0378 & 0.0244 & \second{0.0622} & 0.0323 & 0.0378 & 0.0239 & 0.0585 & 0.0306 \\
\cellcolor{white}\categoryspan{catTrad}{26pt}{7.4pt}{Traditional\\Recommendation KD} & \rule[-2.5pt]{0pt}{9pt}RCE-KD & 0.0388 & 0.0244 & \second{0.0631} & \second{0.0324} & 0.0374 & \second{0.0246} & 0.0620 & \second{0.0325} & 0.0365 & 0.0233 & 0.0587 & 0.0302 \\
\cellcolor{white} & \rule[-2.5pt]{0pt}{9pt}GKD & 0.0372 & 0.0237 & 0.0628 & 0.0317 & \second{0.0382} & 0.0240 & 0.0612 & 0.0317 & 0.0337 & 0.0217 & 0.0544 & 0.0282 \\
\cellcolor{white}\categoryspan{catGen}{17pt}{2.9pt}{General AR /\\LLM KD} & \rule[-2.5pt]{0pt}{9pt}DistiLLM-2 & 0.0376 & 0.0239 & 0.0613 & 0.0313 & 0.0372 & 0.0236 & 0.0579 & 0.0303 & 0.0324 & 0.0208 & 0.0529 & 0.0270 \\
\cellcolor{white} & \rule[-2.5pt]{0pt}{9pt}SID-MLP (Base KD) & 0.0374 & 0.0235 & 0.0626 & 0.0315 & 0.0367 & 0.0233 & 0.0565 & 0.0296 & 0.0328 & 0.0210 & 0.0533 & 0.0276 \\
\cellcolor{white} & \rule[-2.5pt]{0pt}{9pt}SmartGR & 0.0381 & 0.0241 & 0.0622 & 0.0321 & 0.0380 & 0.0238 & 0.0604 & 0.0315 & 0.0330 & 0.0206 & 0.0519 & 0.0267 \\
\cellcolor{white}\categoryspan{catSID}{26pt}{7.4pt}{SID-based generative\\recommendation KD} & \rule[-2.5pt]{0pt}{9pt}\redcell{\textbf{MuSID-KD}} & \redcell{\textbf{0.0408}} & \redcell{\textbf{0.0254}} & \redcell{\textbf{0.0644}} & \redcell{\textbf{0.0330}} & \redcell{\textbf{0.0388}} & \redcell{\textbf{0.0250}} & \redcell{\textbf{0.0650}} & \redcell{\textbf{0.0334}} & \redcell{\textbf{0.0398}} & \redcell{\textbf{0.0250}} & \redcell{\textbf{0.0639}} & \redcell{\textbf{0.0328}} \\
\cmidrule(l){2-14}
\cellcolor{white} & \textit{Gain vs. best (\%)} & +2.77\% & +3.25\% & +2.06\% & +1.85\% & +1.57\% & +1.63\% & +4.50\% & +2.77\% & +4.74\% & +3.73\% & +5.97\% & +5.13\% \\
\cellcolor{white} & \textit{Gain vs. Base (\%)} & +9.09\% & +8.09\% & +2.88\% & +4.76\% & +5.72\% & +7.30\% & +15.04\% & +12.84\% & +21.34\% & +19.05\% & +19.89\% & +18.84\% \\
\cellcolor{white} & SD & 0.00028 & 0.00011 & 0.00039 & 0.00008 & 0.00011 & 0.00008 & 0.00022 & 0.00017 & 0.00022 & 0.00008 & 0.00022 & 0.00014 \\
\midrule
\multicolumn{14}{@{}l}{\textbf{VG}} \\
\cellcolor{white}\catrect{secondCell}{12pt}{Teacher reference} & \rule[-2.5pt]{0pt}{9pt}TIGER (Teacher) & 0.0612 & 0.0403 & 0.0951 & 0.0512 & 0.0612 & 0.0403 & 0.0951 & 0.0512 & 0.0612 & 0.0403 & 0.0951 & 0.0512 \\
\cellcolor{white} & \rule[-2.5pt]{0pt}{9pt}RD & 0.0575 & 0.0372 & 0.0926 & 0.0484 & 0.0589 & 0.0381 & 0.0933 & 0.0493 & 0.0580 & 0.0374 & \second{0.0921} & 0.0483 \\
\cellcolor{white} & \rule[-2.5pt]{0pt}{9pt}RRD & 0.0582 & \second{0.0390} & 0.0918 & 0.0495 & 0.0594 & 0.0387 & 0.0893 & 0.0490 & \second{0.0591} & \second{0.0390} & 0.0884 & 0.0485 \\
\cellcolor{white}\categoryspan{catTrad}{26pt}{7.4pt}{Traditional\\Recommendation KD} & \rule[-2.5pt]{0pt}{9pt}RCE-KD & 0.0586 & 0.0388 & 0.0920 & \second{0.0497} & 0.0587 & \second{0.0394} & 0.0937 & \second{0.0504} & 0.0589 & 0.0388 & 0.0918 & \second{0.0494} \\
\cellcolor{white} & \rule[-2.5pt]{0pt}{9pt}GKD & 0.0590 & 0.0381 & 0.0928 & 0.0493 & \second{0.0598} & 0.0392 & 0.0931 & 0.0497 & 0.0570 & 0.0370 & 0.0897 & 0.0475 \\
\cellcolor{white}\categoryspan{catGen}{17pt}{2.9pt}{General AR /\\LLM KD} & \rule[-2.5pt]{0pt}{9pt}DistiLLM-2 & \second{0.0592} & 0.0379 & 0.0924 & 0.0486 & 0.0596 & 0.0385 & \second{0.0939} & 0.0502 & 0.0587 & 0.0383 & 0.0915 & 0.0489 \\
\cellcolor{white} & \rule[-2.5pt]{0pt}{9pt}SID-MLP (Base KD) & 0.0588 & 0.0383 & 0.0922 & 0.0488 & 0.0591 & 0.0389 & 0.0935 & 0.0499 & 0.0553 & 0.0358 & 0.0870 & 0.0460 \\
\cellcolor{white} & \rule[-2.5pt]{0pt}{9pt}SmartGR & 0.0584 & 0.0386 & \second{0.0930} & 0.0491 & 0.0585 & 0.0383 & 0.0929 & 0.0495 & 0.0550 & 0.0360 & 0.0872 & 0.0462 \\
\cellcolor{white}\categoryspan{catSID}{26pt}{7.4pt}{SID-based generative\\recommendation KD} & \rule[-2.5pt]{0pt}{9pt}\redcell{\textbf{MuSID-KD}} & \redcell{\textbf{0.0601}} & \redcell{\textbf{0.0396}} & \redcell{\textbf{0.0944}} & \redcell{\textbf{0.0505}} & \redcell{\textbf{0.0607}} & \redcell{\textbf{0.0400}} & \redcell{\textbf{0.0954}} & \redcell{\textbf{0.0512}} & \redcell{\textbf{0.0605}} & \redcell{\textbf{0.0399}} & \redcell{\textbf{0.0944}} & \redcell{\textbf{0.0508}} \\
\cmidrule(l){2-14}
\cellcolor{white} & \textit{Gain vs. best (\%)} & +1.52\% & +1.54\% & +1.51\% & +1.61\% & +1.51\% & +1.52\% & +1.60\% & +1.59\% & +2.37\% & +2.31\% & +2.50\% & +2.83\% \\
\cellcolor{white} & \textit{Gain vs. Base (\%)} & +2.21\% & +3.39\% & +2.39\% & +3.48\% & +2.71\% & +2.83\% & +2.03\% & +2.61\% & +9.40\% & +11.45\% & +8.51\% & +10.43\% \\
\cellcolor{white} & SD & 0.00011 & 0.00014 & 0.00034 & 0.00011 & 0.00034 & 0.00017 & 0.00064 & 0.00025 & 0.00031 & 0.00025 & 0.00045 & 0.00031 \\
\bottomrule
\end{tabular*}
\endgroup
\end{table*}

\begin{table*}[!t]
\centering
\caption{Amazon ablations (R@10 and N@10). HCDA-only and RIGD-only isolate each loss while retaining cross entropy. Bold: MuSID-KD; gray: strongest competing variant.}
\label{tab:ablation}
\begingroup\fontsize{6.5}{7.9}\selectfont
\setlength{\tabcolsep}{1.05pt}\renewcommand{\arraystretch}{0.98}
\begin{tabular*}{\textwidth}{@{\extracolsep{\fill}}l*{18}{c}@{}}
\toprule
 & \multicolumn{6}{c}{\textbf{Beauty}} & \multicolumn{6}{c}{\textbf{Toys}} & \multicolumn{6}{c}{\textbf{Video Games}}\\
\cmidrule(lr){2-7}\cmidrule(lr){8-13}\cmidrule(l){14-19}
\textbf{Variant} & \multicolumn{2}{c}{MLP} & \multicolumn{2}{c}{GRU} & \multicolumn{2}{c}{CNN} & \multicolumn{2}{c}{MLP} & \multicolumn{2}{c}{GRU} & \multicolumn{2}{c}{CNN} & \multicolumn{2}{c}{MLP} & \multicolumn{2}{c}{GRU} & \multicolumn{2}{c}{CNN}\\
 & R & N & R & N & R & N & R & N & R & N & R & N & R & N & R & N & R & N\\
\midrule
CE-only & 0.0660 & 0.0359 & 0.0388 & 0.0215 & 0.0288 & 0.0161 & 0.0591 & 0.0303 & 0.0333 & 0.0189 & 0.0261 & 0.0140 & 0.0925 & 0.0492 & 0.0646 & 0.0342 & 0.0375 & 0.0196 \\
Base KD & 0.0660 & 0.0356 & 0.0656 & 0.0358 & 0.0580 & 0.0319 & 0.0626 & 0.0315 & 0.0565 & 0.0296 & 0.0533 & 0.0276 & 0.0922 & 0.0488 & 0.0935 & 0.0499 & 0.0870 & 0.0460 \\
\addlinespace[1pt]
HCDA-only & 0.0666 & 0.0361 & 0.0666 & 0.0361 & 0.0609 & 0.0335 & 0.0631 & 0.0321 & 0.0594 & 0.0309 & 0.0565 & 0.0291 & 0.0929 & 0.0493 & 0.0939 & 0.0503 & 0.0895 & 0.0477 \\
RIGD-only & 0.0667 & 0.0360 & 0.0667 & 0.0362 & 0.0617 & 0.0336 & 0.0632 & 0.0320 & 0.0596 & 0.0308 & 0.0573 & 0.0296 & 0.0928 & 0.0495 & 0.0939 & 0.0504 & 0.0894 & 0.0476 \\
\addlinespace[1pt]
Leaf-only & 0.0666 & 0.0360 & 0.0664 & 0.0361 & 0.0613 & 0.0332 & 0.0631 & 0.0319 & 0.0591 & 0.0307 & 0.0566 & 0.0294 & 0.0928 & 0.0491 & 0.0941 & 0.0503 & 0.0892 & 0.0474 \\
Leaf+Pair & 0.0668 & 0.0363 & 0.0667 & \second{0.0363} & 0.0619 & 0.0337 & 0.0633 & 0.0320 & 0.0596 & 0.0310 & 0.0568 & 0.0293 & 0.0931 & 0.0495 & 0.0942 & 0.0504 & 0.0896 & 0.0478 \\
Leaf+RIGD & \second{0.0671} & \second{0.0364} & 0.0669 & 0.0362 & \second{0.0634} & 0.0342 & 0.0636 & 0.0321 & 0.0602 & 0.0314 & 0.0581 & \second{0.0302} & 0.0929 & 0.0494 & 0.0943 & 0.0506 & 0.0900 & 0.0479 \\
\addlinespace[1pt]
HCDA+Uniform & 0.0665 & 0.0363 & \second{0.0670} & 0.0359 & 0.0626 & 0.0340 & 0.0634 & \second{0.0324} & 0.0599 & 0.0313 & 0.0576 & 0.0297 & 0.0926 & 0.0496 & 0.0936 & 0.0504 & \second{0.0904} & 0.0482 \\
HCDA+AbsRes & 0.0670 & 0.0362 & 0.0668 & \second{0.0363} & 0.0632 & \second{0.0344} & 0.0635 & 0.0321 & \second{0.0608} & \second{0.0316} & \second{0.0582} & 0.0301 & \second{0.0933} & \second{0.0498} & \second{0.0946} & 0.0505 & 0.0902 & \second{0.0484} \\
HCDA+PosUni & 0.0669 & 0.0362 & 0.0666 & 0.0362 & 0.0621 & 0.0339 & \second{0.0637} & 0.0322 & 0.0603 & 0.0311 & 0.0577 & 0.0299 & 0.0930 & 0.0497 & 0.0944 & \second{0.0507} & 0.0903 & 0.0483 \\
HCDA+Shuffled & 0.0667 & 0.0361 & 0.0665 & 0.0361 & 0.0620 & 0.0338 & 0.0632 & 0.0323 & 0.0600 & 0.0312 & 0.0579 & 0.0298 & 0.0932 & 0.0493 & 0.0945 & 0.0501 & 0.0901 & 0.0480 \\
HCDA+Collapsed & 0.0664 & 0.0358 & 0.0662 & 0.0360 & 0.0602 & 0.0331 & 0.0629 & 0.0318 & 0.0581 & 0.0305 & 0.0561 & 0.0290 & 0.0927 & 0.0490 & 0.0940 & 0.0502 & 0.0887 & 0.0471 \\
\redcell{\textbf{MuSID-KD}} & \redcell{\textbf{0.0679}} & \redcell{\textbf{0.0369}} & \redcell{\textbf{0.0682}} & \redcell{\textbf{0.0368}} & \redcell{\textbf{0.0686}} & \redcell{\textbf{0.0368}} & \redcell{\textbf{0.0644}} & \redcell{\textbf{0.0330}} & \redcell{\textbf{0.0650}} & \redcell{\textbf{0.0334}} & \redcell{\textbf{0.0639}} & \redcell{\textbf{0.0328}} & \redcell{\textbf{0.0944}} & \redcell{\textbf{0.0505}} & \redcell{\textbf{0.0954}} & \redcell{\textbf{0.0512}} & \redcell{\textbf{0.0944}} & \redcell{\textbf{0.0508}} \\
\bottomrule
\end{tabular*}
\endgroup
\end{table*}

\begin{figure*}[!t]
\centering\includegraphics[width=.94\textwidth]{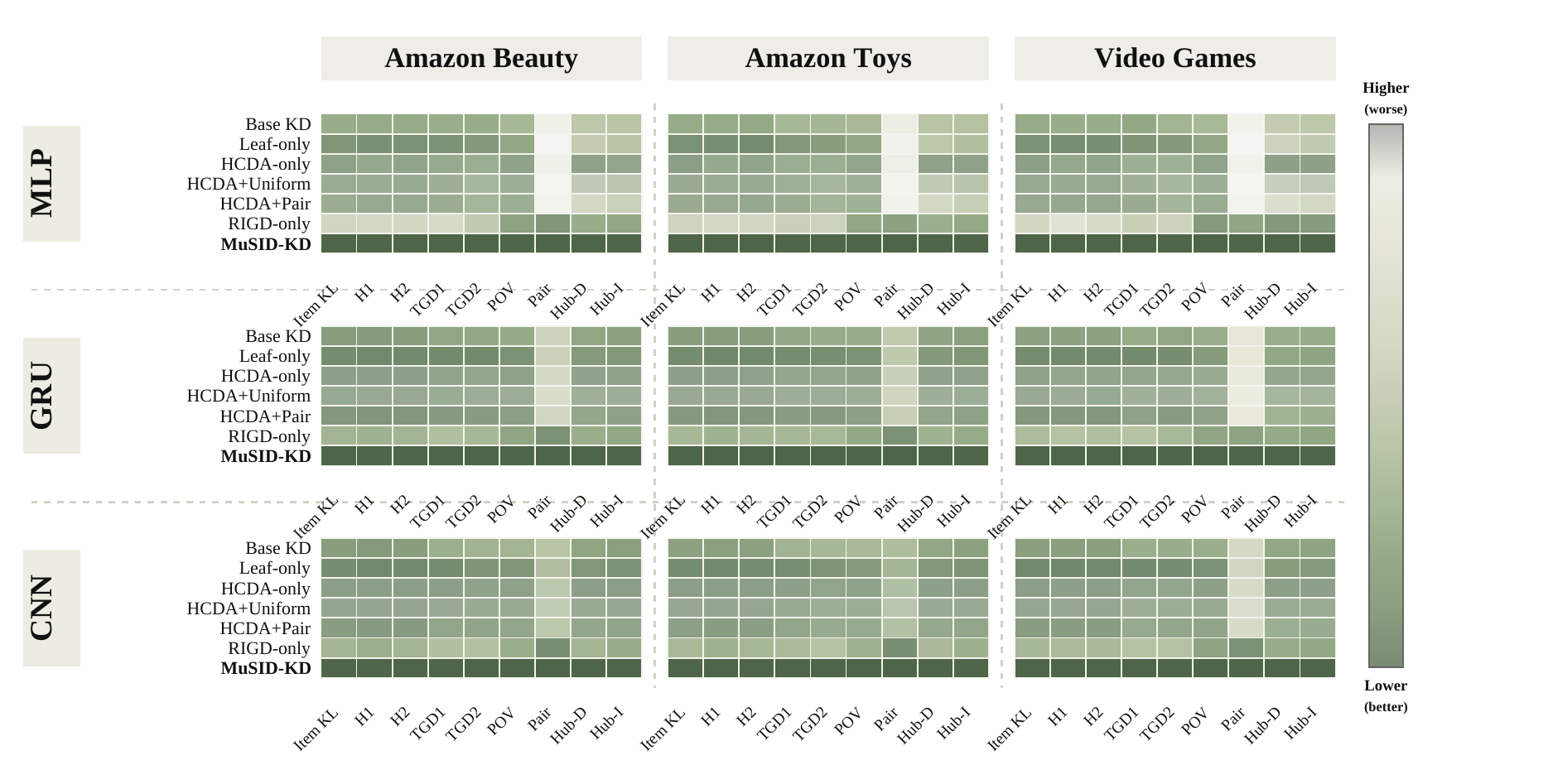}
\Description{Nine heatmaps across Beauty, Toys, and Video Games and MLP, GRU, and CNN, with seven training objectives and nine diagnostic metrics each. Darker colors indicate lower error within each dataset, student, and metric.}
\caption{Nine diagnostics of candidate distribution alignment and item ranking across Amazon Beauty, Toys, and Video Games with MLP, GRU, and CNN. H1/H2 denote HDrift-1/2; TGD1/TGD2 denote TGD-1/2; Pair denotes Pair Softplus; and Hub-D/Hub-I denote Huber-D/I. Each metric is normalized within each dataset and student setting; darker colors indicate lower error.}
\label{fig:diagnostics}
\end{figure*}

\subsection{Experimental Settings}
\noindent\textbf{Datasets and evaluation.} Our main experiments use Amazon Beauty, Toys, and Video Games, supplemented by Steam-100K. Beauty and Toys use Amazon Reviews 2014~\citep{amazon2014}, while Video Games uses Amazon Reviews 2023~\citep{amazon2023}. We sort interactions chronologically and use leave-one-out evaluation for next item prediction; each request includes up to 20 preceding interactions. All Amazon and Steam-100K experiments are conducted with three independent random seeds (42, 43, and 44). We report the mean performance across the three runs. All methods follow the same data splits and evaluation protocol, including identical item mappings, SID assignments, and valid SID constraints. Recall@5/10 and normalized discounted cumulative gain (NDCG)@5/10 measure performance against held out interactions. NDCG@10 is the primary metric. MuSID-KD's Amazon metric standard deviations are in Table~\ref{table:main}.

\noindent\textbf{Teacher, students, and baselines.} We use a fixed TIGER teacher and three lightweight student architectures: an attention readout (MLP), recurrent readout (GRU), and temporal convolution readout (CNN). Each student uses four prefix-conditioned prediction heads and shares the teacher's frozen history encoder, item token mapping, token embeddings, and valid SID constraints. All methods use constrained beam search with width 50. We compare conventional recommendation KD (RD, RRD, RCE-KD), general generative distillation (GKD, DistiLLM-2), and SID specific distillation (SID-MLP, SmartGR). We define \emph{Base KD} as SID-MLP's token distribution KL combined with cross entropy at ground-truth prefixes. Importantly, Base KD is not the same as \emph{Leaf-only}, which aligns teacher and student distributions over full candidate items. Other distillation methods are adapted to the same teacher, student architectures, and decoding protocol. Appendix~\ref{app:implementation} details the implementations.

\noindent\textbf{Implementation.} Each training request uses up to $M=64$ teacher candidate items. RIGD samples pairs with teacher ranks $1$ through $10$ versus $11$ through $40$ ($K=10$, $B=40$). The learning rate is $5\times10^{-5}$; checkpoints are selected using validation NDCG@10, without consulting test performance. Training hyperparameters are specified in Appendix~\ref{app:implementation}.

\subsection{Overall Recommendation Performance (RQ1)}
Table~\ref{table:main} reports results for three student architectures on the three Amazon datasets. MuSID-KD achieves the highest mean among student methods in all \textbf{36} comparisons (three datasets, three students, four metrics). Its mean relative NDCG@10 improvement over Base KD is \textbf{8.31\%} across the nine settings. Relative to the strongest competing student method in each setting, NDCG@10 improves by \textbf{1.59\%--5.13\%}. The improvement holds for MLP, GRU, and CNN students, indicating that the proposed objectives are not restricted to one lightweight architecture. On Steam-100K, NDCG@10 improves over Base KD by \textbf{2.15\%} on average across architectures, with complete results in Appendix~\ref{app:steam}.

\begin{figure}[!t]
\centering
\includegraphics[width=.91\columnwidth,trim=0 .36in 0 0,clip]{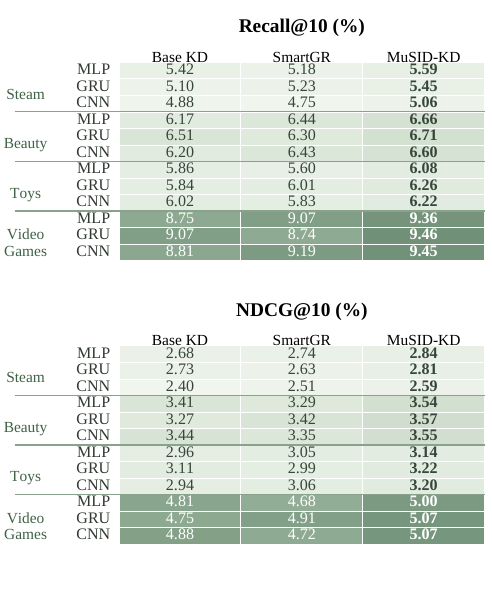}
\Description{Two heatmaps of Recall at 10 and NDCG at 10 on Steam, Beauty, Toys, and Video Games for MLP, GRU, and CNN students. Columns are Base KD, SmartGR, and MuSID-KD. MuSID-KD is highest in all rows for both metrics.}
\caption{Results using residual K-means (R-KMeans) SIDs instead of the main RQ-VAE tokenizer. Teacher and students use the same R-KMeans SIDs. Values are Recall@10 and NDCG@10 in percent; bold marks the best student per dataset and architecture.}
\label{fig:rkmeans}
\end{figure}

\noindent\textbf{Generalization across SID tokenizers.} The preceding experiments use SIDs constructed with a residual quantization variational autoencoder (RQ-VAE). To test whether the proposed distillation objectives depend on this particular tokenizer, we additionally evaluate SIDs generated by residual K-means (R-KMeans). The teacher and student models in this setting use the same R-KMeans SID assignment. As shown in Figure~\ref{fig:rkmeans}, MuSID-KD outperforms both Base KD and SmartGR on Recall@10 and NDCG@10 for all four datasets and all three student architectures, and ranks first in all \textbf{24 settings}. Its NDCG@10 improves by \textbf{5.50\%} on average over Base KD, and by \textbf{3.61\%} over the stronger of Base KD and SmartGR in each setting. Together with the RQ-VAE results in Table~\ref{table:main}, these findings indicate that MuSID-KD's benefits are not restricted to one SID tokenization method.

\subsection{Ablation Study (RQ2)}
Table~\ref{tab:ablation} evaluates R@10/N@10 ablations; the @5 and Steam-100K results appear in Appendix~\ref{app:additional-results}. MuSID-KD variants share ground truth cross entropy; HCDA-only and RIGD-only isolate each proposed loss, and the full method combines them. Base KD additionally includes token distribution KL.

\noindent\textbf{Contribution of HCDA.} Leaf-only supervises the final candidate item distribution; Leaf+RIGD also includes RIGD while omitting intermediate candidate group alignment. Full MuSID-KD improves the mean NDCG@10 over Leaf+RIGD by \textbf{4.21\%} across the nine Amazon settings. The HCDA+Shuffled variant deliberately breaks the correspondence between candidate probabilities and their SID prefixes; its lower performance supports the importance of the correct candidate group structure rather than simply adding more KL terms.

\noindent\textbf{Contribution of RIGD.} HCDA+Uniform weights eligible SID levels equally; HCDA+AbsRes uses absolute teacher--student score residuals, and HCDA+PosUni gives equal weights only to positive residuals. Full MuSID-KD outperforms all three alternatives in NDCG@10 across the nine settings. The absolute residual alone measures correction size rather than improvement to the item order. PosUni selects directionally useful positions without distinguishing how much each helps. Leaf+Pair directly optimizes the full pairwise score gap and remains below full MuSID-KD. HCDA+Collapsed aligns the combined residual, which also changes normalization; the comparison therefore does not isolate supervision granularity on its own. Together, these results favor selection and weighting by replacement gain at each SID position.

\subsection{Knowledge Transfer Analysis (RQ3)}
We evaluate nine diagnostics on the teacher's top 40 candidate items for each request. \emph{Item KL} measures divergence between candidate item distributions. \emph{HDrift-1/2} measure divergence between prefix marginal distributions at SID positions 1 and 2, while \emph{TGD-1/2} measure disagreement on the most probable prefix group. For ranking, \emph{POV} is the rate at which the student reverses or ties item pairs ranked by the teacher; \emph{Pair Softplus} is the mean pairwise ranking surrogate loss; and \emph{Huber-D/I} measure errors in pairwise score contributions under two fixed weighting rules derived from Base KD. All measures are lower when alignment is better; precise definitions, units, and raw values are in Appendix~\ref{app:mechanism}.

Figure~\ref{fig:diagnostics} shows on Beauty that HCDA-only reduces intermediate group discrepancies beyond Leaf-only. On Beauty with CNN, HDrift-2 decreases from 0.0663 to 0.0386 and TGD-2 from 31.58\% to 25.23\%. This indicates better alignment of group probability mass and the teacher's most probable prefix group. When HCDA is kept fixed, MuSID-KD lowers Pair Softplus, POV, and Huber errors relative to HCDA+Uniform. For Beauty with MLP, Pair Softplus drops from 0.3883 to 0.3414. The results support complementary gains in distribution alignment across SID granularities and final item ordering.

\subsection{Training and Serving Efficiency (RQ4)}
All additional computation in MuSID-KD is incurred during training: building group distributions, generating item pairs spanning the teacher's top $K$ cutoff, performing replacements at individual SID positions, and determining weights. On Beauty with MLP, 200 training epochs including beam search validation take 27 minutes for Base KD versus 49 minutes for MuSID-KD. Peak GPU memory grows from 7.4 GB to 9.2--9.5 GB. Inference does not compute distillation targets. For MLP, GRU, and CNN, Base KD/MuSID-KD throughputs are 2769/3036, 3445/3423, and 3028/3024 requests/s, respectively, indicating comparable serving efficiency. Further cost breakdowns are in Appendix~\ref{app:cost}.

\section{Conclusion and Future Work}
MuSID-KD addresses two limitations in SID distillation: incomplete transfer of rankings among candidate groups sharing SID prefixes and different ranking effects of teacher alignment across SID positions. HCDA aligns candidate group distributions, while RIGD weights position-level score alignment by its impact on pairwise ranking loss. Across datasets and student architectures, the method improves accuracy and knowledge transfer without changing online decoding. Future work will consider efficient candidate sampling, ranking impact estimation, and joint corrections across SID positions.

\FloatBarrier
\nocite{*}
\bibliographystyle{ACM-Reference-Format}
\bibliography{refs}

\providecommand{\catrect}[3]{%
  \colorbox{#1}{\parbox[c]{72pt}{\fontsize{6.7}{7.6}\selectfont\bfseries #3}}}
\providecommand{\ablatag}[2]{%
  \colorbox{#1}{\makebox[88pt][l]{\fontsize{6.8}{7.8}\selectfont #2}}}
\providecommand{\second}[1]{\colorbox{secondCell}{#1}}
\definecolor{catTrad}{HTML}{E7F1FC}
\definecolor{catGen}{HTML}{FFF2DC}
\definecolor{catSID}{HTML}{E9F3E6}
\definecolor{serveRow}{HTML}{F5DCD8}
\definecolor{bestRow}{HTML}{F5DCD8}
\definecolor{secondCell}{HTML}{ECEDEF}
\definecolor{ablHServeTag}{HTML}{DCEAF8}
\definecolor{ablDATag}{HTML}{F9E9C9}
\definecolor{ablRule}{HTML}{DDE2E9}

\appendix
\setcounter{figure}{0}
\renewcommand{\thefigure}{A\arabic{figure}}
\section{Additional Experimental Results}
\label{app:additional-results}

\begin{table*}[!t]
\centering
\renewcommand{\thetable}{S1}
\caption{Steam-100K baseline comparison (\%). TIGER is a teacher reference and is excluded from student best/second-best highlighting. Bold and gray indicate the best and second-best student results, including ties.}
\label{tab:steam-baselines}
\begingroup
\fontsize{7.0}{8.6}\selectfont
\setlength{\tabcolsep}{1.1pt}
\renewcommand{\arraystretch}{1.10}
\begin{tabular*}{\textwidth}{@{\extracolsep{\fill}}lll*{12}{c}@{}}
\toprule
\textbf{Dataset} & \textbf{Category} & \textbf{Method} & \multicolumn{4}{c}{\textbf{MLP}} & \multicolumn{4}{c}{\textbf{GRU}} & \multicolumn{4}{c}{\textbf{CNN}} \\
\cmidrule(lr){4-7}\cmidrule(lr){8-11}\cmidrule(l){12-15}
 & & & R@5 & N@5 & R@10 & N@10 & R@5 & N@5 & R@10 & N@10 & R@5 & N@5 & R@10 & N@10 \\
\midrule
\cellcolor{white} & \cellcolor{white}\catrect{secondCell}{12pt}{Teacher} & TIGER (Teacher) & 4.25 & 2.43 & 6.15 & 3.03 & 4.25 & 2.43 & 6.15 & 3.03 & 4.25 & 2.43 & 6.15 & 3.03 \\
\addlinespace[0.5pt]
\cellcolor{white} & \cellcolor{white} & RD & \second{4.17} & 2.36 & 5.59 & \second{2.86} & 3.93 & 2.32 & 5.58 & 2.82 & 3.68 & 2.14 & 5.34 & 2.64 \\
\cellcolor{white} & \cellcolor{white} & RRD & 4.15 & \second{2.39} & 5.65 & 2.84 & \second{4.05} & \second{2.34} & 5.54 & \second{2.83} & 3.72 & 2.13 & 5.32 & 2.64 \\
\cellcolor{white} & \cellcolor{white}\categoryspan{catTrad}{28.7220pt}{11.0866pt}{Traditional\\Recommendation KD} & RCE-KD & 4.01 & 2.33 & 5.61 & 2.83 & 3.92 & 2.32 & 5.50 & 2.78 & 3.74 & \second{2.17} & 5.40 & 2.65 \\
\addlinespace[0.5pt]
\cellcolor{white} & \cellcolor{white} & GKD & 3.82 & 2.26 & 5.62 & 2.80 & 3.81 & 2.32 & 5.48 & 2.81 & 3.69 & 2.12 & 5.38 & 2.62 \\
\cellcolor{white} & \cellcolor{white}\categoryspan{catGen}{18.1190pt}{5.7462pt}{General AR\\/ LLM KD} & DistiLLM-2 & 3.76 & 2.23 & \second{5.66} & 2.82 & 3.69 & 2.24 & \second{5.59} & 2.79 & \second{3.75} & 2.13 & \second{5.42} & 2.68 \\
\addlinespace[0.5pt]
\cellcolor{white} & \cellcolor{white} & SID-MLP (base KD) & 3.80 & 2.25 & \second{5.66} & 2.84 & 3.72 & 2.27 & 5.49 & \second{2.83} & 3.45 & 2.08 & 5.31 & \second{2.69} \\
\cellcolor{white} & \cellcolor{white} & SmartGR & 3.81 & 2.26 & 5.65 & 2.83 & 3.77 & 2.29 & 5.55 & 2.81 & 3.70 & 2.11 & 5.29 & 2.61 \\
\cellcolor{white}\multirow{-9}{*}{\shortstack[l]{\textbf{Steam-}\\\textbf{100K}}} & \cellcolor{white}\categoryspan{catSID}{27.0910pt}{10.2652pt}{SID-based\\generative KD} & \redcell{\textbf{MuSID-KD}} & \redcell{\textbf{4.26}} & \redcell{\textbf{2.43}} & \redcell{\textbf{5.77}} & \redcell{\textbf{2.91}} & \redcell{\textbf{4.19}} & \redcell{\textbf{2.40}} & \redcell{\textbf{5.69}} & \redcell{\textbf{2.88}} & \redcell{\textbf{3.82}} & \redcell{\textbf{2.21}} & \redcell{\textbf{5.51}} & \redcell{\textbf{2.75}} \\
\bottomrule
\end{tabular*}
\endgroup
\end{table*}

\subsection{Steam-100K}
\label{app:steam}
Table~\ref{tab:steam-baselines} reports the four recommendation metrics and teacher reference. Table~\ref{tab:steam-ablation} reports the 14 ablation configurations on the same metrics. Table~\ref{tab:amazon-ablation-at5} presents Amazon @5 ablation results complementary to the @10 results in the main text.


\subsection{Hyperparameter Sensitivity}
\label{app:empirical-sensitivity}
\begin{figure}[H]
\centering
\includegraphics[width=.96\columnwidth]{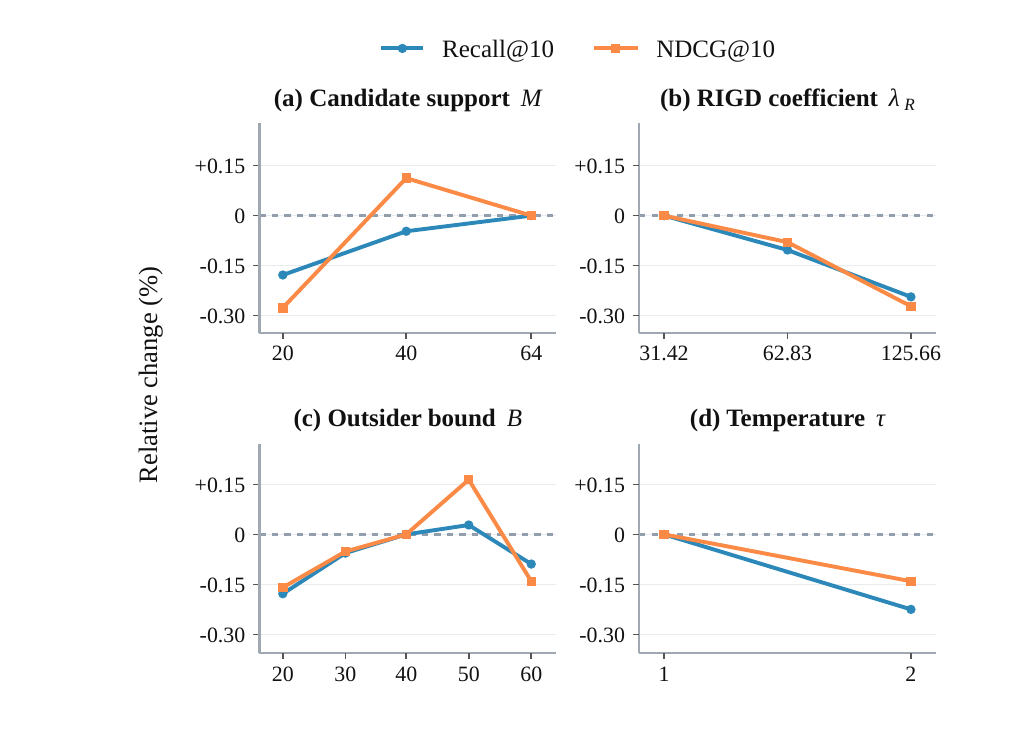}
\Description{Four line charts of Recall at 10 and NDCG at 10 against candidate count M, RIGD weight lambda R, rank cutoff B, and temperature tau.}
\caption{Beauty--MLP hyperparameter sensitivity: relative Recall@10 and NDCG@10 gains (\%) over the MuSID-KD reference result; dashed horizontal lines mark zero relative change.}
\label{fig:empirical-sensitivity}
\end{figure}

Figure~\ref{fig:empirical-sensitivity} reports Beauty--MLP sensitivity to $M$, $\lambda_R$, $B$, and $\tau$. Rel. gain (\%) is the percentage change in Recall@10 or NDCG@10 relative to the MuSID-KD reference result for the corresponding sweep. The dashed horizontal lines mark zero relative change; the default parameter values are $M=64$, $\lambda_R=31.42$, $B=40$, and $\tau=1$. For candidate lists truncated below $B$, the effective pair boundary is $B_{\mathrm{eff}}=\min(B,|\mathcal C_M(u)|)$, with $K<B_{\mathrm{eff}}\le |\mathcal C_M(u)|$.

\begin{table*}[!t]
\centering
\renewcommand{\thetable}{S2}
\caption{Steam-100K ablations on all four metrics (\%). Base KD and MuSID-KD provide reference results for the ablation comparisons. Bold indicates best student results and gray denotes second-best (ties included).}
\label{tab:steam-ablation}
\begingroup
\fontsize{6.65}{7.9}\selectfont
\setlength{\tabcolsep}{1.25pt}
\renewcommand{\arraystretch}{1.05}
\begin{tabular*}{\textwidth}{@{\extracolsep{\fill}}l*{12}{c}@{}}
\toprule
\textbf{Setting} & \multicolumn{4}{c}{\textbf{MLP}} & \multicolumn{4}{c}{\textbf{GRU}} & \multicolumn{4}{c}{\textbf{CNN}} \\
\cmidrule(lr){2-5}\cmidrule(lr){6-9}\cmidrule(l){10-13}
 & R@5 & N@5 & R@10 & N@10 & R@5 & N@5 & R@10 & N@10 & R@5 & N@5 & R@10 & N@10 \\
\midrule
CE-only & 3.62 & 2.26 & 5.54 & 2.82 & 3.68 & 2.27 & 5.46 & 2.79 & 3.51 & 2.11 & 5.39 & 2.69 \\
Base KD & 3.80 & 2.25 & 5.66 & 2.84 & 3.72 & 2.27 & 5.49 & \second{2.83} & 3.45 & 2.08 & 5.31 & 2.69 \\
\arrayrulecolor{ablRule}\midrule\arrayrulecolor{black}
\ablatag{ablHServeTag}{Leaf-only} & 4.03 & 2.35 & 5.55 & \second{2.86} & 3.93 & 2.30 & \second{5.60} & 2.82 & 3.71 & 2.13 & 5.28 & 2.67 \\
\ablatag{ablHServeTag}{Leaf+Pair} & 4.11 & \second{2.38} & 5.56 & 2.80 & 3.92 & 2.30 & 5.51 & 2.76 & 3.64 & 2.12 & \second{5.41} & \second{2.70} \\
\ablatag{ablHServeTag}{Leaf+RIGD} & 4.12 & 2.37 & 5.63 & 2.85 & 4.10 & 2.33 & 5.52 & 2.75 & 3.71 & 2.15 & 5.37 & 2.63 \\
\ablatag{ablHServeTag}{HCDA-only} & 4.17 & 2.34 & 5.57 & 2.79 & 4.09 & \second{2.36} & 5.56 & 2.80 & \second{3.76} & 2.10 & 5.36 & 2.63 \\
\arrayrulecolor{ablRule}\midrule\arrayrulecolor{black}
\ablatag{ablDATag}{HCDA+Pair} & 4.18 & 2.35 & 5.64 & 2.81 & 4.07 & 2.35 & 5.57 & 2.79 & 3.65 & 2.14 & 5.33 & \second{2.70} \\
\ablatag{ablDATag}{RIGD-only} & 4.13 & \second{2.38} & 5.60 & 2.81 & 3.88 & 2.28 & 5.47 & 2.76 & 3.67 & 2.15 & 5.38 & 2.67 \\
\ablatag{ablDATag}{HCDA+Uniform} & 4.16 & 2.34 & 5.64 & 2.85 & 4.08 & 2.31 & 5.58 & 2.75 & \second{3.76} & 2.15 & 5.39 & 2.66 \\
\ablatag{ablDATag}{HCDA+AbsRes} & \second{4.19} & 2.32 & \second{5.68} & 2.78 & 4.11 & 2.35 & 5.55 & 2.79 & 3.73 & 2.16 & 5.37 & 2.68 \\
\ablatag{ablDATag}{HCDA+PosUni} & 4.14 & 2.36 & 5.58 & 2.78 & 4.06 & 2.33 & 5.53 & 2.77 & 3.72 & 2.16 & \second{5.41} & 2.65 \\
\ablatag{ablDATag}{HCDA+Shuffled} & 4.10 & 2.37 & 5.67 & 2.83 & \second{4.12} & \second{2.36} & 5.57 & 2.81 & 3.73 & \second{2.17} & 5.30 & 2.63 \\
\ablatag{ablDATag}{HCDA+Collapsed} & 3.56 & 2.16 & 5.23 & 2.68 & 3.34 & 2.06 & 5.09 & 2.62 & 3.70 & 2.10 & 5.35 & 2.66 \\
\arrayrulecolor{ablRule}\midrule\arrayrulecolor{black}
\redcell{\textbf{MuSID-KD (Full)}} & \redcell{\textbf{4.26}} & \redcell{\textbf{2.43}} & \redcell{\textbf{5.77}} & \redcell{\textbf{2.91}} & \redcell{\textbf{4.19}} & \redcell{\textbf{2.40}} & \redcell{\textbf{5.69}} & \redcell{\textbf{2.88}} & \redcell{\textbf{3.82}} & \redcell{\textbf{2.21}} & \redcell{\textbf{5.51}} & \redcell{\textbf{2.75}} \\
\bottomrule
\end{tabular*}
\endgroup
\end{table*}

\begin{table*}[!t]
\centering
\renewcommand{\thetable}{S3}
\caption{Complementary Amazon ablations at @5 (\%). Each entry reports \emph{Recall@5 / NDCG@5}; the corresponding @10 results are in the main text. Bold denotes MuSID-KD.}
\label{tab:amazon-ablation-at5}
\begingroup
\fontsize{7.0}{8.5}\selectfont
\setlength{\tabcolsep}{2.1pt}
\renewcommand{\arraystretch}{1.04}
\begin{tabular*}{\textwidth}{@{\extracolsep{\fill}}l*{9}{c}@{}}
\toprule
\textbf{Setting} & \multicolumn{3}{c}{\textbf{Beauty}} & \multicolumn{3}{c}{\textbf{Toys}} & \multicolumn{3}{c}{\textbf{Video Games}} \\
\cmidrule(lr){2-4}\cmidrule(lr){5-7}\cmidrule(l){8-10}
 & MLP & GRU & CNN & MLP & GRU & CNN & MLP & GRU & CNN \\
\midrule
CE-only & 4.34/2.87 & 2.53/1.72 & 1.78/1.26 & 3.70/2.32 & 2.19/1.52 & 1.72/1.11 & 5.87/3.83 & 4.05/2.65 & 2.32/1.50 \\
Base KD & 4.29/2.82 & 4.16/2.81 & 3.80/2.54 & 3.74/2.35 & 3.67/2.33 & 3.28/2.10 & 5.88/3.83 & 5.91/3.89 & 5.53/3.58 \\
\addlinespace[1.2pt]
\cellcolor{ablHServeTag}Leaf-only & 4.31/2.85 & 4.23/2.83 & 3.97/2.65 & 3.82/2.41 & 3.73/2.36 & 3.52/2.22 & 5.91/3.87 & 5.93/3.93 & 5.70/3.71 \\
\cellcolor{ablHServeTag}HCDA-only & 4.34/2.83 & 4.24/2.84 & 3.96/2.63 & 3.85/2.43 & 3.73/2.37 & 3.51/2.21 & 5.90/3.87 & 5.97/3.90 & 5.71/3.71 \\
\cellcolor{ablHServeTag}Leaf+Pair & 4.35/2.83 & 4.25/2.85 & 4.05/2.67 & 3.85/2.42 & 3.75/2.42 & 3.54/2.23 & 5.93/3.88 & 5.96/3.90 & 5.72/3.73 \\
\cellcolor{ablHServeTag}Leaf+RIGD & 4.36/2.85 & 4.24/2.85 & 4.07/2.72 & 3.90/2.43 & 3.76/2.40 & 3.61/2.30 & 5.94/3.90 & 5.99/3.93 & 5.77/3.76 \\
\cellcolor{ablHServeTag}HCDA+Shuffled & 4.35/2.86 & 4.26/2.84 & 4.04/2.66 & 3.88/2.39 & 3.77/2.43 & 3.56/2.25 & 5.92/3.85 & 5.97/3.95 & 5.75/3.75 \\
\addlinespace[1.2pt]
\cellcolor{ablDATag}RIGD-only & 4.32/2.87 & 4.26/2.85 & 3.99/2.66 & 3.87/2.42 & 3.75/2.38 & 3.50/2.24 & 5.91/3.89 & 5.96/3.94 & 5.72/3.74 \\
\cellcolor{ablDATag}HCDA+Uniform & 4.33/2.86 & 4.28/2.86 & 4.02/2.69 & 3.91/2.45 & 3.78/2.39 & 3.55/2.27 & 5.90/3.88 & 5.94/3.91 & 5.78/3.77 \\
\cellcolor{ablDATag}HCDA+AbsRes & 4.34/2.84 & 4.27/2.86 & 4.08/2.70 & 3.87/2.44 & 3.80/2.41 & 3.58/2.28 & 5.96/3.91 & 6.00/3.94 & 5.76/3.78 \\
\cellcolor{ablDATag}HCDA+PosUni & 4.37/2.87 & 4.29/2.82 & 4.03/2.68 & 3.89/2.46 & 3.79/2.38 & 3.59/2.26 & 5.95/3.89 & 5.98/3.94 & 5.74/3.74 \\
\cellcolor{ablDATag}HCDA+Collapsed & 4.32/2.84 & 4.21/2.84 & 3.94/2.61 & 3.81/2.40 & 3.71/2.37 & 3.44/2.21 & 5.89/3.86 & 5.95/3.92 & 5.67/3.69 \\
\redcell{MuSID-KD (Full)} & \redcell{4.42/2.92} & \redcell{4.41/2.91} & \redcell{4.37/2.88} & \redcell{4.08/2.54} & \redcell{3.88/2.50} & \redcell{3.98/2.50} & \redcell{6.01/3.96} & \redcell{6.07/4.00} & \redcell{6.05/3.99} \\
\bottomrule
\end{tabular*}
\endgroup
\end{table*}

\section{Implementation and Measurement Details}
\label{app:implementation}
\subsection{Data, Models, and Training}
Beauty and Toys use Amazon Reviews 2014~\citep{amazon2014}; Video Games uses Amazon Reviews 2023~\citep{amazon2023}. Chronological leave-one-out evaluation uses up to 20 preceding interactions. The frozen TIGER T5 history encoder is shared. MLP, GRU, and CNN students use attention, recurrent, and temporal-convolution/pooling readouts, respectively, followed by four prefix-conditioned MLP heads. Under the default student configurations, readout and prediction-head parameter counts are 1.664M (MLP), 1.763M (GRU), and 1.713M (CNN), excluding frozen encoder parameters and token embeddings. Beam width is $W=50$.

The learning rate is $5\times10^{-5}$. Training uses $M=64$ teacher candidates and pairs between ranks 1--10 and 11--40 ($K=10$, $B=40$). HCDA uses $\tau=1$, $\omega_d=1$, and a $10^{-9}$ group-probability floor. RIGD uses $\mu=0.05$, $\epsilon=10^{-9}$, and unit-threshold Huber loss. Its scale is $\sigma_d=\max\{\operatorname{Std}[g_{T,d}],10^{-4}\}$, estimated from teacher pairwise score contributions at eligible SID positions in the training data. Pairs without positive gain remain in the averaging denominator with zero contribution; an empty pair set gives zero loss.

The effective coefficients are $\lambda_H=4$ and $\lambda_R=31.42$. HCDA is averaged over requests and RIGD over valid minibatch pairs. The additional objectives, caches, and scale statistics use training requests only. Amazon and Steam-100K experiments use three independent seeds (42, 43, and 44); the reported performance is averaged across runs. Steam-100K ablations use 200 training epochs.

Base KD combines $0.3$ token cross-entropy and $0.7$ token KL at ground-truth prefixes. Adapted external objectives retain this base. RD uses rank-decayed preferences and dynamic student ranks; RRD samples catalog negatives; RCE-KD uses ranking cross-entropy. GKD mixes ground-truth and sampled prefixes, and DistiLLM-2 uses skew divergences on contrastive sampled SIDs. Their teacher targets are detached and queried conditionals are normalized over the selected 256-token codebook. SmartGR combines token and beam-prefix objectives. CE-only omits distillation. Leaf-only retains only candidate item distribution KL; Leaf+Pair and Leaf+RIGD add the indicated supervision. The explicit direct-Pair controls use $\mathcal L_{\mathrm{Pair}}=|\mathcal B|^{-1}\sum_{(i,j)\in\mathcal B}\operatorname{softplus}(-G_S(i,j)/0.05)$ with coefficient one, rather than RIGD's Huber-weighted depth correction.

\paragraph{Ablation implementations.}
HCDA+Uniform weights eligible nonfinal SID levels equally. For a pair $(i,j)$, let $r_d=g_{T,d}-g_{S,d}$ at eligible level $d$. HCDA+AbsRes first gates on whether any $r_d>0$ and then normalizes $|r_d|$ over eligible levels; consequently, SID positions with negative score residuals can still receive nonzero weights. HCDA+PosUni assigns equal weights only where $r_d>0$, setting all weights to zero when there is no positive residual. HCDA+Shuffled applies the same within-request permutation to teacher and student candidate score vectors while keeping SID leaf positions fixed, disrupting the correspondence between probabilities and prefix groups. HCDA+Collapsed sums eligible residuals before applying one Huber penalty, normalized using the current minibatch standard deviation of the collapsed residual rather than per-level fixed scales. Thus, Collapsed changes both supervision granularity and the normalization strategy. HCDA+Pair and Leaf+Pair directly optimize the direct pairwise ranking loss above, unlike RIGD's replacement gain weighted alignment of score contributions.

\subsection{Scoring and Mechanism Diagnostics}
\label{app:mechanism}
Diagnostics use the teacher's top 40 candidates per request (training uses $M=64$). Teacher and student item probabilities $\bar\pi_T$ and $\bar\pi_S$ normalize their respective cached item and student SID-path scores on this shared support. Item KL is $\mathrm{KL}(\bar\pi_T\Vert\bar\pi_S)$. HDrift-$d$ is the KL between length-$d$ prefix marginals, each obtained by summing item probabilities; depths 1 and 2 are reported to avoid redundancy with item KL. Top Group Disagreement (TGD-$d$) flags requests where the most probable length-$d$ prefix differs between teacher and student; SID-prefix ties are broken lexicographically.

\begin{table*}[!t]
\centering
\renewcommand{\thetable}{B1}
\caption{Training, GPU memory, cache export, and serving costs. Times are minutes or ms/request (serving); training includes beam validation.}
\label{tab:current-cost}
\begingroup\footnotesize
\setlength{\tabcolsep}{2pt}
\begin{minipage}[t]{0.33\textwidth}
\centering\textbf{(a) Full training and memory}\par\smallskip
\begin{tabular*}{\linewidth}{@{\extracolsep{\fill}}lrr@{}}
\toprule
Run & \shortstack{200 epochs\\min} & \shortstack{Peak GPU\\GB}\\
\midrule
MLP Base KD & 27 & 7.4\\
\redcell{MLP MuSID-KD} & \redcell{49} & \redcell{9.2--9.5}\\
CNN MuSID-KD & 65 & --\\
\bottomrule
\end{tabular*}
\end{minipage}\hfill
\begin{minipage}[t]{0.26\textwidth}
\centering\textbf{(b) Teacher cache export}\par\smallskip
\begin{tabular*}{\linewidth}{@{\extracolsep{\fill}}lr@{}}
\toprule Dataset & min\\\midrule
Beauty & 26.2\\Toys & 26.7\\Video Games & 38.6\\
\bottomrule\end{tabular*}
\end{minipage}\hfill
\begin{minipage}[t]{0.36\textwidth}
\centering\textbf{(c) MuSID-KD serving phases}\par\smallskip
\begin{tabular*}{\linewidth}{@{\extracolsep{\fill}}lrr@{}}
\toprule Phase & GRU & CNN\\\midrule
Frozen T5 encoder & 0.154 & 0.151\\
Student context & 0.031 & 0.046\\
SID prediction/beam scoring & 0.089 & 0.114\\
Beam bookkeeping & 0.019 & 0.019\\\midrule
Total & 0.292 & 0.331\\\bottomrule
\end{tabular*}
\end{minipage}
\endgroup
\end{table*}

\begin{table*}[!t]
\begingroup
\centering
\renewcommand{\thetable}{D1}
\caption{Beauty diagnostics (lower is better). TGD and POV are percentages; other metrics are unscaled. Bold marks the minimum; definitions are in Appendix~\ref{app:mechanism}.}
\label{tab:mechanism-all}
\footnotesize
\setlength{\tabcolsep}{2.2pt}
\renewcommand{\arraystretch}{0.92}
\begin{tabular*}{\textwidth}{@{\extracolsep{\fill}}lllrrrrrrrrr@{}}
\toprule
Dataset & Student & Method & Item KL & HDrift-1 & HDrift-2 & TGD-1 & TGD-2 & POV & \shortstack{Pair\\Softplus} & Huber-D & Huber-I \\
\midrule
\multirow{21}{*}{Beauty} & \multirow{7}{*}{MLP} & Base KD & 0.100670 & 0.045033 & 0.082126 & 16.79 & 35.75 & 10.93 & 0.393655 & 0.076414 & 0.053440 \\
 &  & Leaf-only & 0.051456 & 0.021539 & 0.042083 & 12.46 & 26.80 & 7.26 & 0.391118 & 0.050309 & 0.032503 \\
 &  & HCDA-only & 0.045387 & 0.018498 & 0.036893 & 11.83 & 25.88 & 6.76 & 0.395989 & 0.051863 & 0.032359 \\
 &  & HCDA+Uniform & 0.044023 & 0.019050 & 0.036987 & 11.93 & 26.05 & 6.33 & 0.388316 & 0.037520 & 0.023753 \\
 &  & HCDA+Pair & 0.045503 & 0.018693 & 0.037182 & 11.89 & 26.35 & 6.54 & 0.395344 & 0.054633 & 0.033297 \\
 &  & RIGD-only & 0.070771 & 0.031490 & 0.058785 & 14.42 & 29.17 & 6.68 & 0.343984 & 0.042486 & 0.026364 \\
\cellcolor{white} & \cellcolor{white} & \redcell{MuSID-KD} & \redcell{\textbf{0.042977}} & \redcell{\textbf{0.017830}} & \redcell{\textbf{0.035638}} & \redcell{\textbf{11.66}} & \redcell{\textbf{25.32}} & \redcell{\textbf{6.23}} & \redcell{\textbf{0.341443}} & \redcell{\textbf{0.036775}} & \redcell{\textbf{0.022948}} \\
\addlinespace[1pt]
 & \multirow{7}{*}{GRU} & Base KD & 0.244564 & 0.119998 & 0.208229 & 26.31 & 50.64 & 17.77 & 0.439444 & 0.151225 & 0.111759 \\
 &  & Leaf-only & 0.048274 & 0.020718 & 0.040485 & 11.11 & 24.86 & 6.51 & 0.375833 & 0.036761 & 0.022435 \\
 &  & HCDA-only & 0.035289 & 0.014068 & 0.028949 & 9.35 & 21.58 & 5.19 & 0.374753 & 0.030595 & 0.018436 \\
 &  & HCDA+Uniform & 0.033592 & 0.014219 & 0.028497 & 9.44 & 21.86 & 4.88 & 0.371142 & 0.022807 & 0.014068 \\
 &  & HCDA+Pair & 0.035688 & 0.014302 & 0.029282 & 9.58 & 22.10 & 5.22 & 0.375735 & 0.031373 & 0.018830 \\
 &  & RIGD-only & 0.073914 & 0.032750 & 0.062797 & 13.65 & 27.54 & 6.14 & 0.327674 & 0.042099 & 0.025052 \\
\cellcolor{white} & \cellcolor{white} & \redcell{MuSID-KD} & \redcell{\textbf{0.032992}} & \redcell{\textbf{0.013668}} & \redcell{\textbf{0.027935}} & \redcell{\textbf{9.20}} & \redcell{\textbf{21.29}} & \redcell{\textbf{4.76}} & \redcell{\textbf{0.324326}} & \redcell{\textbf{0.022391}} & \redcell{\textbf{0.013709}} \\
\addlinespace[1pt]
 & \multirow{7}{*}{CNN} & Base KD & 0.515006 & 0.318404 & 0.462264 & 42.81 & 66.64 & 24.69 & 0.538513 & 0.262580 & 0.199730 \\
 &  & Leaf-only & 0.077514 & 0.037424 & 0.066335 & 16.03 & 31.58 & 9.40 & 0.388830 & 0.051249 & 0.031182 \\
 &  & HCDA-only & 0.046750 & 0.019938 & 0.038603 & 11.28 & 25.23 & 6.59 & 0.380494 & 0.037289 & 0.022323 \\
 &  & HCDA+Uniform & 0.042607 & 0.019081 & 0.036282 & 11.12 & 24.06 & 5.97 & 0.375686 & 0.027136 & 0.016653 \\
 &  & HCDA+Pair & 0.047297 & 0.020728 & 0.039247 & 12.09 & 25.15 & 6.63 & 0.380145 & 0.038250 & 0.022788 \\
 &  & RIGD-only & 0.136456 & 0.067351 & 0.116790 & 19.30 & 35.68 & 8.76 & 0.328085 & 0.072418 & 0.043099 \\
\cellcolor{white} & \cellcolor{white} & \redcell{MuSID-KD} & \redcell{\textbf{0.041376}} & \redcell{\textbf{0.018824}} & \redcell{\textbf{0.035351}} & \redcell{\textbf{10.89}} & \redcell{\textbf{23.63}} & \redcell{\textbf{5.80}} & \redcell{\textbf{0.323154}} & \redcell{\textbf{0.026011}} & \redcell{\textbf{0.016244}} \\
\bottomrule
\end{tabular*}
\endgroup
\end{table*}

For item pairs spanning the teacher's top $K$ cutoff, ranks 1--10 are paired with ranks 11--40. Write $G_S(i,j)=\sum_{d=1}^L g_{S,d}(i,j)$, consistent with Eq.~\eqref{eq:globalgap}. Pairwise Order Violation (POV) counts a student score gap $G_S\le0$, including ties, as a violation; Pair Softplus averages $\operatorname{softplus}(0.05-G_S)$, which differs from the \emph{training control} Pair loss $\operatorname{softplus}(-G_S/0.05)$. These diagnostics use common pairs and are not conditional on a positive replacement gain.

Huber-D and Huber-I use identical per-level normalized residuals $z_d=(g_{S,d}-g_{T,d})/\sigma_d$ and the unit-threshold Smooth L1 penalty $\rho_H(z_d)$ over eligible levels $D_{ij}$ (from the SID divergence depth through level 3). Their weights are determined once using a \emph{fixed Base KD reference}, not separately optimized for each evaluated model. Specifically, for $q\in\{D,I\}$, define the reference replacement gain at a single SID position and its normalized nonnegative weight:
\begin{equation}
\begin{aligned}
L_D(G)&=\operatorname{softplus}(0.05-G),\\
L_I(G)&=\operatorname{softplus}(-G/0.05),\\
a_d^{(q)}&=[L_q(G_{\rm base})-L_q(G_{\rm base}+r_{d,\rm base})]_+,\\
w_d^{(q)}&=a_d^{(q)}/\max\{\textstyle\sum_{t\in D_{ij}}a_t^{(q)},10^{-9}\}.
\end{aligned}
\label{eq:diagnostic-huber-weights}
\end{equation}
Huber-$q$ averages $\sum_{d\in D_{ij}}w_d^{(q)}\rho_H(z_d)$ over valid pairs. Both are diagnostics of score contributions at individual SID positions; neither applies Huber directly to the pairwise score gap. Item KL, HDrift, and TGD are averaged by request; POV, Pair Softplus, and both Huber variants are averaged across all valid item pairs. TGD and POV are displayed as percentages, while all KL, Softplus, and Huber values are unscaled. Relative percentage changes of recommendation scores are averaged over dataset--student settings after computing $(v_{\mathrm{MuSID\text{-}KD}}/v_{\mathrm{control}}-1)\times100\%$ for each setting.
\subsection{Additional Cost Measurements}
\label{app:cost}
Table~\ref{tab:current-cost} reports training, memory, caching, and serving costs. Student serving uses batch size 32; teacher throughput uses 64.

\section{Objective Properties}
\label{app:properties}
By the KL chain rule, HCDA induces position weights $(4,3,2,1)$ on conditional distributions over SID prefixes when $L=4$ and $\omega_d=1$, compared with $(1,1,1,1)$ for Leaf-only. For a teacher ranked pair, a positive replacement gain at position $d$ is equivalent to a positive score residual $r_d=g_{T,d}-g_{S,d}$. The residual and the replacement gain are different quantities: the former is a score difference, whereas the latter is a reduction in pairwise ranking surrogate loss.

\section{Exact Mechanism Measurements}
\label{app:mechanism-data}
Table~\ref{tab:mechanism-all} gives Beauty diagnostic measurements for MLP, GRU, and CNN; see Appendix~\ref{app:mechanism} for definitions and units.

\paragraph{Consistency on the other datasets.}
All nine metrics use the same evaluation rules on Toys and Video Games. MuSID-KD consistently improves over Base KD for MLP, GRU, and CNN. For Toys--MLP, Item KL, TGD-2, and POV decrease from 0.139783, 39.02\%, and 13.62\% to 0.060499, 28.02\%, and 8.01\%, respectively; for Video Games--MLP, they decrease from 0.076808, 31.93\%, and 9.12\% to 0.031806, 21.24\%, and 4.74\%.

\end{document}